\documentclass[10pt,twocolumn,letterpaper]{article}

\usepackage{cvpr}      

\usepackage{amsfonts} 
\usepackage[accsupp]{axessibility}
\usepackage{dblfloatfix}
\usepackage{nicefrac}       
\usepackage{microtype}      
\usepackage{graphicx}
\usepackage{amsmath}
\usepackage{amssymb}
\usepackage{booktabs}
\usepackage{times}
\usepackage{graphicx}
\usepackage{amsmath}
\usepackage{amssymb}
\usepackage{dsfont}
\usepackage{siunitx}
\usepackage{multirow}
\usepackage{subcaption}
\usepackage{adjustbox}
\usepackage{enumitem}
\usepackage{tablefootnote}
\usepackage{bbm}
\usepackage{xspace}
\usepackage{mathtools}
\usepackage{xcolor}
\usepackage{color, colortbl}
\definecolor{Gray}{gray}{0.9}
\usepackage[normalem]{ulem}

\definecolor{citecolor}{HTML}{0071bc}
\usepackage[pagebackref,breaklinks,colorlinks,citecolor=citecolor]{hyperref}

\usepackage[capitalize]{cleveref}

\crefname{section}{Section}{Sections}
\Crefname{section}{Section}{Sections}
\Crefname{table}{Table}{Tables}
\crefname{table}{Table}{Tables}
\crefname{figure}{Figure}{Figures}
\Crefname{figure}{Figure}{Figures}

\def\cvprPaperID{3375} 
\def\confName{CVPR}
\def\confYear{2022}

\DeclarePairedDelimiter{\norm}{\lVert}{\rVert}

\definecolor{darkgreen}{rgb}{0.0,0.6,0.0}

\DeclareMathOperator*{\argmin}{arg\,min}

\makeatletter
\newcommand{\thickhline}{%
    \noalign {\ifnum 0=`}\fi \hrule height 1pt
    \futurelet \reserved@a \@xhline
}
\newcommand{\reals}{\mathbb{R}}

\newcommand{\bb}{\boldsymbol{b}}
\newcommand{\xx}{\boldsymbol{x}}

\newcommand{\indic}[1]{\mathds{1}_{\{#1\}}}

\newcommand{\zz}{\boldsymbol{z}}

\newcommand{\vv}{\boldsymbol{v}}
\newcommand{\cc}{\boldsymbol{c}}
\newcommand{\pp}{\boldsymbol{p}}

\newcommand{\ignorebig}[1]{}

\newcommand{\normm}[1]{\left\|#1\right\|}
\newcommand{\comment}[1]{}

\newcommand{\model}{DETReg\xspace}

\newcommand{\defdetr}{Deformable~DETR\xspace}

\newcommand{\imagenetthousand}{IN1K\xspace}
\newcommand{\imagenethundred}{IN100\xspace}

\newcommand\minisection[1]{\vspace{1mm}\noindent \textbf{#1}}

\begin{document}


\title{\model: Unsupervised Pretraining with \\Region Priors for Object Detection}
\author{
  \textbf{Amir Bar}$^{1}$, \textbf{Xin Wang}$^{5}$, \textbf{Vadim Kantorov}$^{1}$, \textbf{Colorado J Reed}$^{2}$, \textbf{Roei Herzig}$^{1}$, \\
  \textbf{Gal Chechik}$^{3,4}$, \textbf{Anna Rohrbach}$^{2}$, \textbf{Trevor Darrell}$^{2}$, \textbf{Amir Globerson}$^{1}$\\
  $^1$ Tel-Aviv University ~~~ $^2$  Berkeley AI Research ~~~ $^3$  NVIDIA ~~~       $^4$  Bar-Ilan University ~~~ $^5$ Microsoft Research\\
\texttt{amir.bar@cs.tau.ac.il}}

\maketitle


\begin{abstract}

Recent self-supervised pretraining methods for object detection largely focus on pretraining the backbone of the object detector, neglecting key parts of detection architecture. Instead, we introduce \model, a new self-supervised method that pretrains the entire object detection network, including the object localization and embedding components. During pretraining, \model predicts object localizations to match the localizations from an unsupervised region proposal generator and simultaneously aligns the corresponding feature embeddings with embeddings from a self-supervised image encoder. We implement \model using the DETR family of detectors and show that it improves over competitive baselines when finetuned on COCO, PASCAL VOC, and Airbus Ship benchmarks. In low-data regimes \model achieves improved performance, e.g., when training with only 1\% of the labels and in the few-shot learning settings.\footnote{Code: \url{https://www.amirbar.net/detreg/}.}

\end{abstract}

\section{Introduction}
\begin{figure}
\centering
\resizebox{1\linewidth}{!}{
    \includegraphics[width=.8\textwidth]{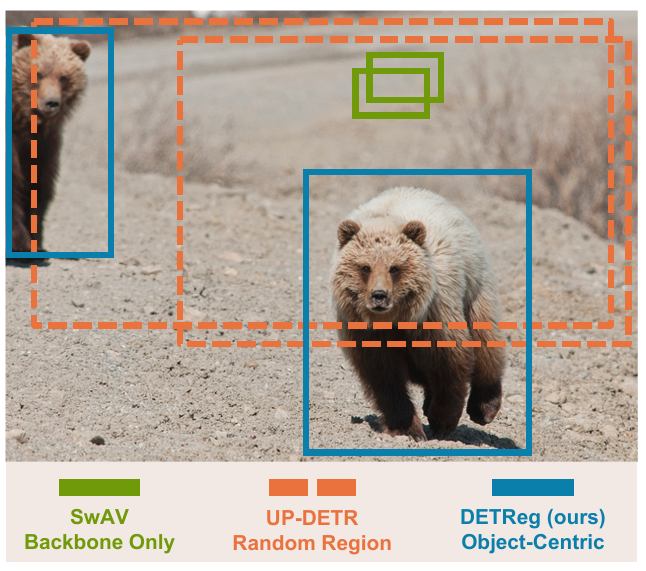}}
    \captionof{figure}{
    \textbf{Top class-agnostic object detections after pretraining.} Self-supervised pretraining methods, such as SwAV~\cite{caron2020unsupervised}, pretrain only the detector's backbone, so the object localizations following the pretraining stage solely depend on the random initialization of the localization components (green).  UP-DETR~\cite{dai2020up} pretrains the entire detection network, but since its pretraining operates by re-identifying random regions, it does not specialize in localizing objects after the pretraining (orange dashes). Our model, \model, pretrains the entire detection network using object-centric pretraining, and following the pretraining stage can localize objects (blue).} 
\label{fig:teaser}
\end{figure}

\begin{figure*}[t]
 \centering
     \centering
     \includegraphics[width=0.95\textwidth]{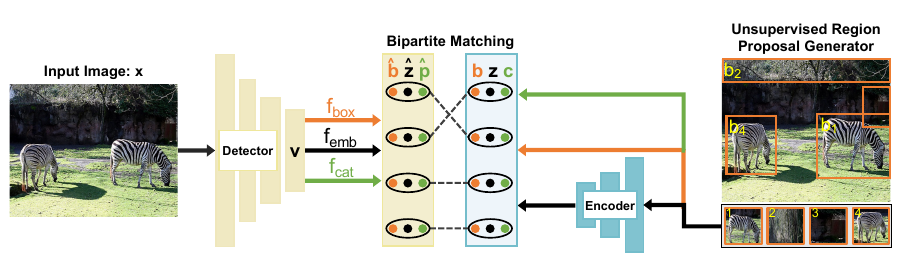}
     \caption{The \textbf{DETReg} model and pretext tasks. Given embeddings $v$ from image $x$, we use the DETR family of detectors~\cite{carion2020end,zhu2020deformable} to predict region proposals ($f_{box}(v)=\hat{b}$), associated object embeddings ($f_{emb}(v)=\hat{z}$), and object scores ($f_{cat}(v) = \hat{p}$). Pseudo ground-truth region proposals labels ($b$) can be generated by existing unsupervised region proposal methods like~\cite{uijlings2013selective,wu2020multi}, and pseudo ground-truth object embeddings (z) can be generated via existing self-supervised approaches like~\cite{he2020momentum, caron2020unsupervised}, where the object score $c$ is always $1$ for these proposals. Predictions are assigned to pseudo labels via Bipartite Matching, and unmatched predictions are assigned with padded proposals with $c=0$.}
    \label{fig:model}
\end{figure*}
Object detection is a key task in computer vision, yet it largely relies on the availability of human-annotated training datasets.
Building such datasets is not only costly but sometimes infeasible for privacy-sensitive applications such as medical imaging or personal photos~\cite{rahimi2021addressing, yu2016iprivacy}. %
Fortunately, recent advancements in self-supervised representation learning have substantially reduced the amount of labeled data needed for a variety of applications, including object detection~\cite{he2020momentum,caron2020unsupervised,grill2020bootstrap,chen2020big}. 

Despite this recent progress, current approaches are limited in their ability to learn good representations for object detection because they do not pretrain the entire object detection network, specifically the localization and region embedding components.
Most recent works (e.g., SwAV~\cite{caron2020unsupervised}, ReSim~\cite{xiao2021region}, InsLoc~\cite{yang2021insloc}) follow the same \textit{pretraining playbook} for the detection network as a supervised image-classification-based pretraining, where only the CNN backbone can be initialized from the pretrained model.
While the recent UP-DETR~\cite{dai2020up} method pretrains a full detection architecture, it still does not localize objects within the image, but rather random image regions.

In this work, we present a model for Detection with Transformers using Region priors (\model), which unlike existing pretraining methods, learns to both localize and encode objects simultaneously in the unsupervised pretraining stage -- see \Cref{fig:teaser}. \model involves two object-centric and category-agnostic pretraining tasks: an \emph{Object Localization Task} to localize objects, and an \emph{Object Embedding Task} to encode an object's visual properties. Taken together, these tasks pretrain the entire detection network -- see \cref{fig:model} for an overview. A final object classification head can then be finetuned with a small number of labels yielding better performance than  existing  methods.

\model's object localization task uses simple region proposal methods for class-agnostic bounding-box supervision~\cite{uijlings2013selective, arbelaez2014multiscale, arbelaez2014multiscale, cheng2014bing, cheng2014global}. 
These methods require little or no training data and can produce region proposals at a high recall. For example, Selective Search~\cite{uijlings2013selective}, the region proposal method we adopt in \model, uses object cues such as continuity in color, hierarchy, and edges to extract object proposals. \model  builds upon these region priors to learn a class-agnostic detector during pretraining.

\model's object embedding task aims to predict the embeddings of a separate self-supervised image encoder evaluated on object regions. 
Self-supervised image encoders, e.g., SwAV~\cite{caron2020unsupervised}, learn transformation-invariant embeddings, so training the detector to predict these values distills the learned invariances into the detector's embeddings.
Thus, the object embedding head learns representations that are robust to transformations such as translation or image cropping.

We conduct an extensive evaluation of \model on standard object detection benchmarks like MS~COCO~\cite{lin2014microsoft} and PASCAL~VOC~\cite{everingham2010pascal}, and on an aerial images dataset, Airbus Ship Detection~\cite{airbus}. We find that \model improves the performance using two state-of-the-art base architectures compared to challenging baselines, especially when small amounts of annotated data are available. 

%
Quantitatively, \model improves over a backbone-only image-classification pretraining baseline by 4 AP points on PASCAL~VOC, 1.6 AP points on MS~COCO, and 1.2 AP points on Airbus Ship Detection. Additionally, \model outperforms pretraining baselines in semi-supervised learning when using 1\% to 10\% of data, and on 10 and 30 shot.
Taken together, these results indicate that pretraining an entire detection network, including region proposal prediction and embedding components, is beneficial and that our specific {\model} model realizes new SOTA performance by taking advantage of this object-centric self-supervised pretraining. 

\section{Related Work}

\minisection{Self-supervised pretraining.} Recent work~\cite{goyal2019scaling, chen2020simple, henaff2021efficient,he2019momentum,he2020momentum,caron2020unsupervised,chen2020improved,misra2020self,gidaris2020learning} has shown that self-supervised pretraining can generate powerful representations for transfer learning, even outperforming its supervised counterparts on challenging vision benchmarks~\cite{wu2018unsupervised, chen2020simple}. 
The learned representations transfer well to image classification but the improvement is less significant for instance-level tasks, such as object detection and instance segmentation~\cite{he2019rethinking,zhou2020cheaper,he2020momentum, reed2021self}.

More recently, a number of works~\cite{roh2021spatially,henaff2021efficient, xiao2021region,xie2021detco} focused on learning backbones that can transfer to object detection. In contrast to these works, we pretrain the entire detection network. As we show, pretraining the backbone with an image-patch-based task does not necessarily empower the model to learn \emph{what and where} an object is, and adding weak supervision from the region priors proves beneficial.
%
%

Our approach is also different from semi-supervised object detection approaches~\cite{jeong2019consistency,sohn2020simple,liu2021unbiased,xu2021end} and few-shot detection approaches~\cite{chen2018lstd,karlinsky2019repmet,kang2019few,yan2019meta,wu2020meta,wu2020multi,fan2020few,xiao2020few,wang2020frustratingly,li2021transformation,fan2021generalized,chen2021should,zhang2021meta} as we initialize the detector from a pretrained \model model without further modifying the architecture. Therefore, these approaches can be viewed as complementary to \model.

\minisection{End-to-end object detection.} Detection with transformers (DETR)~\cite{carion2020end} builds
the first fully end-to-end object detector and eliminates the need for components such as anchor generation and
non-maximum suppression (NMS) post-processing.
This model has quickly gained traction in the machine vision community.
However, the original DETR suffers from slow
convergence and limited sample efficiency. \defdetr~\cite{zhu2020deformable} introduced a deformable attention module to attend to a sparsely sampled small set of prominent key elements, and achieved better performance compared to DETR with reduced training epochs. We therefore use \defdetr as our base detection architecture. 

Both DETR and \defdetr adopt 
the supervised pretrained backbone (ResNet~\cite{he2016deep}) on ImageNet. UP-DETR~\cite{dai2020up} pretrains DETR in a self-supervised way by detecting and reconstructing the random patches from the input image. 
Instead, we additionally adopt region priors from unsupervised region proposal algorithms to provide weak supervision for pretraining, which has an explicit notion of \emph{objects} rather than the random patches used by UP-DETR.

\minisection{Region proposals.} A rich study
of region proposals methods exists in the object detection literature~\cite{alexe2010object,van2011segmentation,carreira2011cpmc,endres2013category, arbelaez2014multiscale,zitnick2014edge,cheng2014bing,krahenbuhl2014geodesic}.
Grouping based method, Selective Search~\cite{van2011segmentation}, and window scoring based approach, Objectness~\cite{alexe2010object} are two early and well known proposal methods, which have been widely adopted and supported in  major software libraries (e.g., OpenCV~\cite{opencv_library}).
Selective Search greedily merges superpixels to generate proposals. Objectness relies on visual cues such as multi-scale saliency, color contrast, edge density and superpixel straddling to identify likely regions. 

While the field has largely shifted to learning-based approaches, the key benefit of these models is that they require little or no training data, and can produce region proposals at a high recall~\cite{uijlings2013selective, arbelaez2014multiscale, arbelaez2014multiscale, cheng2014bing, cheng2014global}. This provides a cheap, albeit noisy, source of supervision. Hosang~\etal~\cite{hosang2014good,hosang2015makes} offer a comprehensive analysis over the various region proposals methods, and Selective Search is among the top performing approaches in terms of recall. 
Here, we seek weak supervision from the region proposals generated by Selective Search, which has been widely adopted and proven successful in the well-known detectors such as R-CNN~\cite{girshick2014rich} and Fast R-CNN~\cite{girshick2015fast}. However, our approach is not limited to Selective Search and can employ other proposal methods.





\section{DETReg}
\label{sec:model}
\model is a self-supervised method to fully pretrain object detectors, including their region localization and embedding components. 
At a high level, \model operates by predicting object localizations that match those from an unsupervised region proposal generator, while simultaneously aligning the corresponding feature embeddings with embeddings from a self-supervised image encoder, see~\cref{fig:model}. 

The key idea underlying \model is to formulate pretext tasks that are similar to the tasks performed during supervised object detection, so that improved pretraining transfers to the object detector. We built \model based on the DETR family of detectors \cite{carion2020end,zhu2020deformable} due to their implementation simplicity and performance, though other architectures can easily plug into \model. Next, we review DETR, and in the following subsections, we present the object localization and embedding pretext tasks that form the core of \model.

\textbf{DETR summary:} DETR detects up to $N$ objects in an image by iteratively applying attention and feed-forward layers over $N$ object query vectors of a transformer decoder and over the input image features. The last layer of the decoder results in $N$ image-dependent query embeddings that are used to predict bounding box coordinates and object categories.
Formally, consider an input image $\xx\in \reals^{H\times W\times3}$. DETR uses $\xx$ to calculate $N$ image-dependent query embeddings  $\vv_1,\ldots,\vv_N$ with $\vv_i\in\reals^d$. This is achieved  by passing the image through a backbone, followed by a transformer, and processing of the query vectors \cite{carion2020end}. Then, two prediction heads are applied to $\vv_i$. The first,  $f_{box}:\reals^d\to\reals^4$, predicts the bounding boxes. The second, $f_{cat}:\reals^d\to\reals^L$, outputs a distribution over $L$ object categories, including a background ``no object'' category.  

\subsection{Object Localization Task} 
\label{ss:object_localization_task}
\model's object localization pretraining task uses simple region proposal methods for class-agnostic bounding-box supervision (see the orange arrows in~\Cref{fig:model}). We use the output from these methods as they require limited or no training data and can produce region proposals at a high recall~\cite{uijlings2013selective, arbelaez2014multiscale, cheng2014bing, cheng2014global}. We use Selective Search~\cite{uijlings2013selective} as the primary region proposal method for training \model as it is widely available in off-the-shelf computer vision libraries and requires no training data. Selective Search uses object cues such as continuity in color and edges to extract object proposals, and \model further builds upon these region priors to learn a class-agnostic detector.

 Region proposal methods take an image and produce a large set of region proposals at a high recall rate, where some of the regions are likely to contain objects. However, they have low precision and do not output category information, see~\cite{hosang2014good,hosang2015makes}.
 Since the content of non-object boxes tends to be more variable than of object boxes, we expect that deep models can be trained to recognize the visual properties of objects even when given noisy labels. 

Thus, the \textit{Object Localization} pretraining task takes a set of $M$ boxes $\bb_1,\ldots,\bb_M$ (where $\bb_i\in\reals^4$) output by an unsupervised region proposal method and optimizes a loss that minimizes the difference between the detector box predictions (the output of the $f_{box}$ MLP) and these $M$ boxes. Similar to DETR, the loss involves matching the predicted boxes and these $M$ boxes, a process we detail in~\Cref{ss:pretraining}.

Common region proposal methods attempt to sort the region proposals such that proposals that are more likely to be objects appear first, however, the number of proposals is typically large, and the ranking is not precise. Therefore, we explore methods to choose the best regions to use during training. We consider three policies for selecting boxes: 
\vspace{0.1cm} 
\\
\noindent\textbf{Top-K} uses the top-$K$ proposals from the algorithm.
\vspace{0.1cm}
\\
\noindent\textbf{Random-K} uses $K$ random proposals, which may yield lower quality proposals but encourages exploration.
\vspace{0.1cm}
\\
\noindent\textbf{Importance Sampling} relies on the region proposal method ranking but also encourages more diverse proposals. Formally, let $\bb_1, \ldots, \bb_n$ be a set of $n$ sorted region proposals, where the $\bb_i$ has rank $i$. Let $X_i$ be a random variable indicating whether we will output the $\bb_i$. We assign the sampling probability for $X_i$ to be: $Pr(X_i =1) \propto -log(i/n).$
To determine if a box should be included, we randomly sample from its respective distribution.

\subsection{Object Embedding Task}
\label{subsec:embedding}
In the supervised training of object detectors, every box is associated with a class category of the object, which is not available in an unsupervised setting.  Therefore, to learn a strong object embedding, we encode each box region $\bb_i$ via a separate encoder network and obtain embeddings $\zz_i$ that are used as a target for the \model embeddings $\hat{\zz_i}$ (see the black arrows in~\Cref{fig:model}).

The separate encoding network that produces $\zz_i$ could be jointly trained by following similar bootstrapping techniques from works such as BYOL~\cite{grill2020bootstrap} or DINO~\cite{caron2021emerging}. However, for training stability and to reduce the convergence time, we leverage a pretrained, self-supervised model whose embeddings are invariant to many image transformations, e.g.~blurring and color distortions. Here we primarily use a SwAV~\cite{caron2020unsupervised} pretrained model as it is one of the strongest performing methods for pretraining image classifiers and has readily available code and pretrained models.

To predict a corresponding object embedding $\hat{\zz_i}$ in the detector, we introduce an additional MLP $f_{emb}:\reals^d\to\reals^d$ that predicts the object embedding $\hat{\zz_i}$ from the corresponding DETR query embedding, $\vv_i$. This encourages $\vv_i$ to capture the information that is useful for category prediction. The loss is the $L_1$ loss between $\hat{\zz_i}$ and $\zz_i$.


\subsection{\model Pretraining}
\label{ss:pretraining}
Here, we formally describe how \model optimizes the localization and embedding tasks during pretraining. 
Assume that our region proposal method returns $M$ object proposals which are used to generate $M$ bounding boxes $\bb_{i}$ and object descriptors $\zz_{i}$ for $i\in\{1,\ldots,M\}$, and let $y_i = (\bb_i, \zz_i)$ with $y=\{y_i\}_{i=1}^{M}$. \model is trained such that its $N$ outputs align with $y$. 

Let $\vv_1,\ldots,\vv_K$ denote the image-dependent query embeddings calculated by DETR (i.e., the output of the last layer of the DETR decoder). 
\model has three prediction heads: $f_{box}$ which outputs predicted bounding boxes, $f_{cat}$ which predicts if the box is object or background, and $f_{emb}$ which reconstructs the object embedding descriptor. Denote these outputs as: $\hat{\bb}_i = f_{box}(\vv_i)$, $\hat{\zz}_i = f_{emb}(\vv_i)$, $\hat{\pp}_i = f_{cat}(\vv_i)$, and define $\hat{y_i}=(\hat{\bb_i}, \hat{\zz_i}, \hat{p_i})$ and $\hat{y} = \{\hat{y_i}\}_{i=1}^{N}$. 

Following DETR training, we assume that the number of DETR queries $N$ is larger than $M$, so we pad $y$ to obtain $N$ tuples, and assign a label $\cc_i \in \{0, 1\}$ to each box in $y$ to indicate whether it is a region proposal ($\cc_i=1$) or padded proposal ($\cc_i=0$); see the green arrows in \cref{fig:model}. With the DETR family of detectors~\cite{carion2020end,zhu2020deformable}, there are no assumptions on the order of the labels or the predictions and therefore we first match the objects of $y$ to the ones in $\hat{y}$ via the Hungarian bipartite matching algorithm~\cite{kuhn1955hungarian}. Specifically, we find the permutation $\sigma$ that minimizes the optimal matching cost between $y$ and $\hat{y}$:
\begin{equation}
    {\sigma} = \argmin_{\sigma\in\Sigma_N} \sum_{i}^{N} {L_{match}({y_i, \hat{y}_{\sigma(i)}})}
\end{equation}
Where $L_{match}$ is the pairwise matching cost matrix as defined in~\cite{carion2020end, zhu2020deformable} and $\Sigma_{N}$ is the set of all permutations over $\{1\ldots N\}$. Using the optimal $\sigma$, we define the loss as:

\begin{equation}
\begin{aligned}
L(y, \hat{y}) = & \sum_{i=1}^N \lambda_{f} L_{class}(\cc_i, \hat{\pp}_{\sigma_{(i)}}) + \\ 
 & \indic{c_i\neq0}(\lambda_{b} L_{box}({\bb_{i}, \hat{\bb}_{\sigma_{(i)}}}) + \\
 & \lambda_{e}L_{emb}(\zz_i, \hat{\zz}_{\sigma(i)}))
\end{aligned}
\end{equation}
Where $L_{class}$ is the class loss, that can be implemented via Cross Entropy Loss or Focal Loss~\cite{lin2017focal}, and $L_{box}$ is based on the the $L_1$ loss and the Generalized Intersection Over Union (GIoU) loss~\cite{rezatofighi2019generalized}. Finally, we define $L_{emb}$ to be the $L_1$ loss:
\begin{equation}
    L_{emb}(\zz_i, \zz_j) = \normm{\zz_i - \hat{\zz}_j}_1
\end{equation}

\section{Experiments}
We first describe the implementation details and datasets used for our experimentation. We then report how \model performs on the object detection tasks when finetuned on full and low-data regimes, including few-shot learning, and semi-supervised learning. Finally, we conclude with ablations, analyses, and visualizations from \model.
\vspace{0.1cm}
\\
\noindent\textbf{Implementation.} Based on the ablations presented in \Cref{ss:analysis}, the default experiment settings are as follows (see the Suppl. for all details). We initialize the ResNet50 backbone of \model with SwAV~\cite{caron2020unsupervised}, which was pretrained with multi-crop views for 800 epochs on \imagenetthousand, and fix it throughout the pretraining stage. In the object embedding branch, $f_{emb}$ and $f_{box}$ are MLPs with $2$ hidden layers of size $256$ followed by a ReLU~\cite{nair2010rectified} nonlinearity. The output sizes of $f_{emb}$ and $f_{box}$ are $512$ and $4$. $f_{cat}$ is implemented as a single fully-connected layer with $2$ outputs. Unless otherwise noted, we use the \model Top-K region selection variant (see \Cref{ss:object_localization_task}) and set $K=30$ proposals per-image.
\vspace{0.1cm}
\\
\noindent\textbf{Datasets.} We use the following datasets: \textbf{ImageNet ILSRVC 2012} (\imagenetthousand) dataset contains $1.2$M images with $1000$ class categories. As done in prior work~\cite{chen2020simple, xiao2021region, caron2020unsupervised, xie2021detco}, we use the unlabeled \imagenetthousand data for pretraining. Similar to other works~\cite{xiao2021region, he2019momentum,he2020momentum}, we use a subset of \imagenetthousand called \textbf{\imagenethundred} that contains $\sim$125K images and $100$ class categories for several ablation studies.
\textbf{MS~COCO}~\cite{lin2014microsoft} is a popular object detection benchmark that contains $121$K labeled images, where objects from 80 object categories are annotated with bounding boxes. 
\textbf{PASCAL~VOC}~\cite{everingham2010pascal} contains $\sim$20K natural images where the objects from 21 classes are annotated. 
To explore a dataset with different visual properties than the typical object-centric benchmarks, we use the \textbf{Airbus Ship Detection} dataset~\cite{airbus}, which contains $\sim$231K satellite images annotated with bounding boxes of ships. Following~\cite{nie2020attention}, we convert the segmentation masks to bounding boxes and use a 42.5K image subset, with 3K test/val splits.

\begin{table}[t]
\centering
\resizebox{1.\linewidth}{!}{
\begin{tabular}{lcclll}
\toprule
Pretraining & Detector & Epochs & AP   & $\textrm{AP}_{50}$ & $\textrm{AP}_{75}$\\
\midrule\midrule
Supervised & \multirow{4}{*}{DETR} & \multirow{4}{*}{150} & 39.5 & 60.3 & 41.4\\
SwAV~\cite{caron2020unsupervised} &  & & 39.7 & 60.3 & 41.7\\
UP-DETR &  & & 40.5 & 60.8 & 42.6 \\
 \textbf{DETReg} &  & & \textbf{41.9}$^{+1.4}$  & \textbf{61.9}$^{+1.1}$ & \textbf{44.1}$^{+1.5}$ \\
\midrule
Supervised & \multirow{4}{*}{DETR} & \multirow{4}{*}{300} & 40.8 & 61.2 & 42.9\\
SwAV~\cite{caron2020unsupervised} &  &  & 42.1 & {63.1} & 44.5\\
UP-DETR &  &  & 42.8 & 63.0 & 45.3\\
\textbf{DETReg} & &  & \textbf{43.7}$^{+0.9}$  & \textbf{63.7}$^{+0.7}$ & \textbf{46.6}$^{+1.3}$ \\
\midrule
Supervised & \multirow{4}{*}{DDETR} & \multirow{4}{*}{50} & 44.5 & 63.6 & 48.7 \\
SwAV~\cite{caron2020unsupervised} &  &  & 45.2 & 64.0 & 49.5\\
UP-DETR &  &  & 44.7 & 63.7 & 48.6 \\
\textbf{DETReg} &  &  & \textbf{45.5}$^{+0.8}$  & \textbf{64.1}$^{+0.4}$ & \textbf{49.9}$^{+1.3}$\\
\bottomrule
\end{tabular}
}
\captionof{table}{\textbf{Object detection results when trained on MS~COCO \texttt{train2017} and evaluated on \texttt{val2017}}. Both DETReg and UP-DETR are pretrained on \imagenetthousand under comparable settings, while supervised and SwAV only pretrain  the backbone of the object detector. We explore both the DETR and \defdetr (DDETR) architectures; for compatibility with prior work, we finetuned the DETR for 150/300 epochs and DDETR for 50 epochs.\vspace{-2mm}}
\label{table:coco}
\end{table}

\subsection{Object Detection in Full Data Regimes}
\label{subsec:fulldata}
These experiments test how well \model performs when a fully annotated dataset is available for finetuning.
\vspace{0.1cm}
\\
\noindent\textbf{Pretraining.} We pretrain two variants of \model based on DETR~\cite{carion2020end} and \defdetr~\cite{zhu2020deformable} detectors for $5$ and $60$ epochs on \imagenetthousand and \imagenethundred, respectively, where the pretraining schedules are set by proportionally adjusting the schedules used in UP-DETR to equate to the more efficient \defdetr schedules~\cite{zhu2020deformable}.
\vspace{0.1cm}
\\
\noindent\textbf{Baselines.} We compare \model to several closely related state-of-the-art pretraining approaches for object detection with transformers: using a SwAV~\cite{caron2020unsupervised} backbone, a fully pretrained UP-DETR~\cite{dai2020up}, and a supervised baseline backbone.
\vspace{0.1cm}
\\
\noindent\textbf{Experiments.} To evaluate \model, we finetuned it on three different datasets: MS~COCO~\cite{lin2014microsoft}, PASCAL~VOC~\cite{everingham2010pascal}, and Airbus Ship Detection~\cite{airbus}. We perform an extensive comparison on MS~COCO and finetune using similar training schedules as previously reported in~\cite{dai2020up, zhu2020deformable}, using \texttt{train2017} for finetuning and \texttt{val2017} for evaluation. On PASCAL~VOC and Airbus we use \model \defdetr based variant, which is faster to train. On PASCAL~VOC we finetune on \texttt{trainval07+12} for $100$ epochs, dropping the learning rate after $70$ epochs and use the \texttt{test07} for evaluation. For Airbus, we finetune for $100$ epochs, dropping the learning rate after $80$ epochs. 
\vspace{0.1cm}
\\
\noindent\textbf{Results.} \Cref{table:coco} shows that \model consistently outperforms other pretraining strategies using both DETR and \defdetr. For example, \model improves the COCO AP score by 1.4 points compared to UP-DETR when trained for 150 epochs, and in fact, outperforms the $300$ epoch supervised variant after $150$ epochs. Interestingly, using \model pretraining with DETR is competitive with supervised \defdetr, which achieves only $0.8$ points (AP) more, despite significant architectural modifications.

\begin{table}[t]
\centering
\small
\resizebox{.9\linewidth}{!}{
\begin{tabular}{lllllll}

\toprule
\multirow{2}{*}{Method} & \multicolumn{3}{c}{PASCAL~VOC} & \multicolumn{3}{c}{Airbus Ship} \\
 & AP & AP$_{50}$ & AP$_{75}$  & AP & AP$_{50}$ & AP$_{75}$\\
\midrule\midrule
Supervised & 59.5  &  82.6 &  65.6 & 79.8 & 95.8 & 89.4 \\ 
SwAV~\cite{caron2020unsupervised} & 61.0 & 83.0 & 68.1 & 78.3 & 95.7 & 88.7  \\
\textbf{\model} & \textbf{63.5} & \textbf{83.3} & \textbf{70.3} & \textbf{81.0}  & 95.9 & \textbf{89.7} \\
\bottomrule                            
\end{tabular} 
}
\caption{\textbf{Object detection finetuned on PASCAL~VOC and Airbus Ship data.} The model is finetuned on PASCAL~VOC \texttt{trainval07+2012} and evaluated on \texttt{test07} (left), and Airbus Ship Detection finetuned on the train split and evaluated on the 3k test images (right). All models are based on \defdetr~\cite{zhu2020deformable}. Bold values indicate an improvement $\geq 0.3$ AP.
}
\label{table:pascal_main}
\end{table}

\Cref{table:pascal_main} shows that \model improves by $2.5$ (AP) points over SwAV on PASCAL~VOC and by $1.2$ (AP) on Airbus. For reference, by using a specialized architecture for ship detection that builds on a ResNet50 backbone, as well as leveraging the pixel-level annotations, \cite{nie2020attention} obtains a box AP score of $76.1$ on this dataset, $4.9$ points lower than \model, which only uses the bounding box annotations.

\subsection{Object Detection in Low-Data Regimes}
\label{sec:low_data}
\begin{figure}[t]
\centering
\resizebox{1\linewidth}{!}{
    \includegraphics[width=\textwidth]{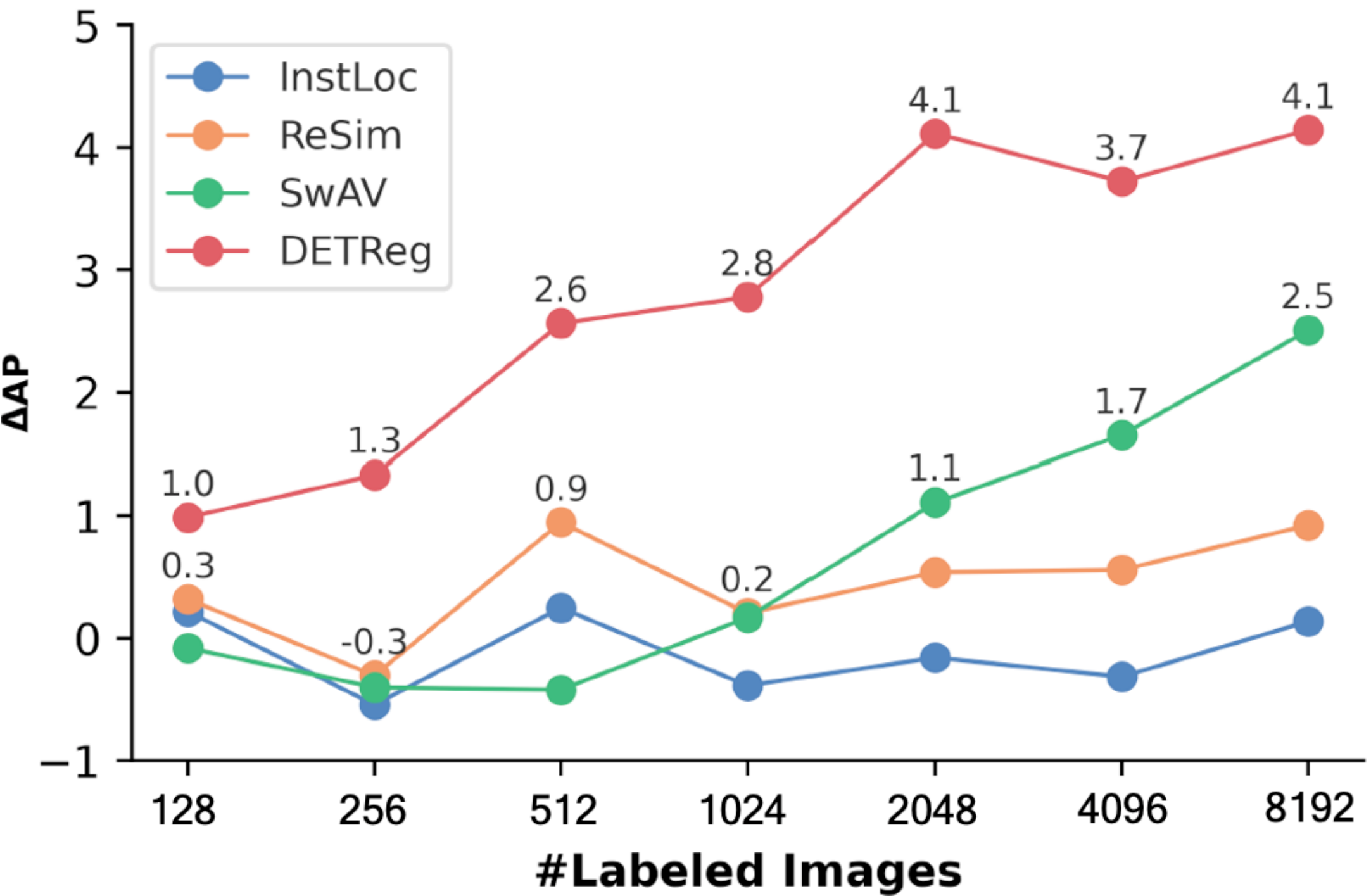}}
    \captionof{figure}{\textbf{Model comparison in low-data regimes.} $\Delta AP$ improvement over the supervised baseline, where the x-axis shows the total number of images used during training. We fix the \defdetr architecture across all methods and finetuned it using publically released ResNet50 weights of different methods on MS~COCO \texttt{train2017} and evaluate on \texttt{val2017}. 
    }

\label{fig:main_coco}
\end{figure}

These experiments test how \model performs when a small amount of annotated data is available for finetuning.
\\
\noindent\textbf{Pretraining.} We pretrain \model based on \defdetr~\cite{zhu2020deformable} for $5$ epochs on ImageNet (\imagenetthousand). 
\\
\noindent\textbf{Baselines.} We consider recent approaches for pretraining ResNet50 backbone for object detection: InstLoc~\cite{yang2021insloc},  ReSim~\cite{xiao2021region} and SwAV~\cite{caron2020unsupervised}, for each we use the publicly released checkpoint. We report the $\Delta AP$ w.r.t. the supervised variant, which utilizes a ResNet50 pretrained on \imagenetthousand.
\\
\noindent\textbf{Experiments.} We test the representations learned by \model when transferring with limited amounts of labeled data (up to $1024$ labeled images), randomly sampled from MS~COCO \texttt{train2017} and use \texttt{val2017} for evaluation. We train all methods for up to $2,000$ epochs or until the validation performance stops improving. 
\vspace{0.1cm}
\\
\noindent\textbf{Results.} \cref{fig:main_coco} shows the results, where the y-axis reports the difference in AP compared to the supervised variant. The results indicate that \model consistently outperforms other pretraining strategies, when using \defdetr on low data regime. For example, when using only $256$ images, \model improves the average precision (AP) score by $4.1$ points compared to $1.1$ for SwAV and $0.5$ for ReSim.

\begin{table}
\makeatletter\def\@captype{table}
    \centering
    \resizebox{.97\linewidth}{!}{
    \begin{tabular}{lcccccc}
    \toprule
    \multirow{2}{*}{Model} 
    & \multirow{2}{*}{Detector} & \multirow{2}{*}{Backbone} &\multicolumn{2}{c}{Novel AP} & \multicolumn{2}{c}{Novel AP$_{75}$}\\
    & & & 10 & 30 & 10 & 30 \\ \midrule\midrule
    \multicolumn{4}{l}{\small\bf\it Methods utilizing small backbones}\\
    \midrule
    YOLO-ft-full\small{~\cite{redmon2017yolo9000,kang2019few}} & \multirow{2}{*}{YOLOv2} & DarkNet-19 & 3.1 & 1.7 & 7.7 & 6.4 \\ 
    FSRW\small{~\cite{kang2019few}} &  & DarkNet-19 & 5.6 & 9.1 & 4.6 & 7.6 \\ 
    \midrule
    FRCN-ft-full\small{~\cite{yan2019meta}} &\multirow{2}{*}{FRCN} & VGG16 & 3.3  &  7.8 & 1.9 & 6.0 \\
    MetaDet\small{~\cite{wang2019meta}} & & VGG16 & 7.1 & 11.3 & 6.1 & 8.1 \\
    \midrule
    FRCN-ft-full\small{~\cite{yan2019meta}} &\multirow{2}{*}{FRCN} & ResNet50 &  6.5 & 11.1 & 5.9 & 10.3\\
    Meta R-CNN\small{~\cite{yan2019meta}} & & ResNet50 & 8.7 & 12.4 & 6.6 & 10.8\\ 
    \midrule
    DDETR-ft-full & \multirow{2}{*}{DDETR} & ResNet50 & 12.4 & 20.4 & 13.3 & 21.8  \\ 
    \textbf{DETReg-ft-full} & & ResNet50 & \textbf{13.7} & \textbf{22.6} & \textbf{15.1} & \textbf{24.3} \\

    \midrule
    \multicolumn{4}{l}{\small\bf\it Methods utilizing large backbones}\\
    \midrule
    FRCN-ft-full~\cite{wang2020frustratingly} &\multirow{3}{*}{FRCN}& ResNet101 & 9.2 & 12.5  & 9.2 & 12.0  \\ 
    TFA~\cite{wang2020frustratingly} & & ResNet101 & {10.0} & {13.7}  & {9.3} & {13.4}\\
    DeFRCN~\cite{zhu2020deformable} & & ResNet101 & {18.5} & \textbf{22.6}  & {-} & {-} \\
    \midrule
    DDETR-ft-full~\cite{zhang2021meta} & \multirow{2}{*}{DDETR*} & ResNet101 & 11.7 & 16.3 & 12.1 & 16.7 \\
    Meta-DETR~\cite{zhang2021meta} & &  ResNet101 &  \textbf{19.0} & {22.2}  & \textbf{19.7} & {22.8}\\
    \midrule
    \textbf{DETReg-ft-full} & DDETR & ResNet50 & 13.7 & \textbf{22.6} & 15.1 & \textbf{24.3} \\
    \bottomrule
    \multicolumn{6}{l}
    \text{\footnotesize \it DDETR$^*$ is the customized single scale  deformable DETR model used in ~\cite{zhang2021meta}}. 
    \end{tabular}
}
\caption{\textbf{Few-shot detection evaluation on COCO.} We trained the model on the 60 base classes and then evaluate the model performance on the 20 novel categories, following the data split used in~\cite{wang2020frustratingly}. Through simple-fine tuning, \model outperforms previous few-shot object detectors with ResNet50 backbones. Using $K=30$, \model achieves similar or improved performance compared to approaches that utilize larger backbones.}
\label{table:fewshot}
\end{table}

\vspace{0.1cm}
\subsection{Few-Shot Object Detection}
These experiments test how \model extends to the few-shot settings established in existing literature.
\vspace{0.1cm}
\\
\noindent\textbf{Pretraining.} We pretrain \model based on \defdetr~\cite{zhu2020deformable} for $5$ epochs on ImageNet (\imagenetthousand). 
\\
\noindent\textbf{Baselines.} We consider \defdetr with a supervised pretrained backbone as the most direct baseline as its architecture and training strategy mirror \model. We also report the results of recent few-shot approaches, which utilize different underlying object detectors. Concurrent to our work, Meta-DETR~\cite{zhang2021meta} proposed a new method based on \defdetr. However, unlike \model, it uses a ResNet101 backbone and a single image scale, but we include its results to encourage unified reporting, even when experimental settings are not perfectly aligned.
\vspace{0.1cm}
\\
\noindent\textbf{Experiments.} Following the standard few-shot protocol for object detection~\cite{wang2020frustratingly}, we finetune \model on the full data of 60 base classes, which contain around $99$K labeled images. Then, we finetune on a balanced set of all 80 classes, where every class has $k\in\{10, 30\}$ object instances. We use the splits from~\cite{wang2020frustratingly} and report the performance on the novel 20 classes. The results are shown in~\cref{table:fewshot}.

\cref{table:fewshot_nobase} shows an extreme few-shot setup where \model is finetuned on the balanced few-shot set without the intermediate finetuning on base classes. We consider finetuning the decoder only (\textit{ft-decoder}) and the full model (\textit{ft-full}). %
\vspace{0.1cm}
\\
\noindent\textbf{Results.} \cref{table:fewshot} shows that \model improves over supervised pretraining and achieves state-of-the-art $30$-shot results compared to approaches that utilize larger backbones. \model only uses a simple fine-tuning strategy, while other methods may include more complicated episodic training. 

\cref{table:fewshot_nobase} shows that \model achieves competitive few-shot performance even when the model is not trained on the abundant base class data. As a reference point, TFA~\cite{wang2020frustratingly} is a previous fine-tuning method that trains on the abundant base class data, and we can see that \model outperforms it without additional supervision from the base class data. 


\begin{table}[t]
\makeatletter\def\@captype{table}
    \centering
    \resizebox{.85\linewidth}{!}{
    \begin{tabular}{lccccc}
    \toprule
    \multirow{2}{*}{Model} 
    & \multirow{2}{*}{Detector} &\multicolumn{2}{c}{Novel AP} & \multicolumn{2}{c}{Novel AP$_{75}$}\\
    & & 10 & 30 & 10 & 30 \\ \midrule\midrule
    TFA~\cite{wang2020frustratingly} (w/base) & FRCN & {10.0} & {13.7}  & {9.3} & {13.4}\\
    \midrule
    DDETR-ft-full & \multirow{4}{*}{DDETR} &  6.7 &  12.5 &  6.0 &  12.4 \\
    DDETR-ft-decoder & & 5.9 &  13.6 & 4.6 & 13.6 \\
    DETReg-ft-decoder & &  7.6 & 15.7 & 7.7 & 16.6 \\
    \textbf{DETReg-ft-full} & & 9.1 & 16.5 & 9.6 & 17.3  \\

    \bottomrule
    \end{tabular}}
\caption{\textbf{Few-shot object detection without training on the COCO base classes.} To test \model's performance on extreme few-shot scenarios, we conduct an evaluation where \model is finetuned only on the K-shot COCO subsets. \model performs slightly worse ($K=10$) or better ($K=30$) compared to TFA~\cite{wang2020frustratingly} without using base class data while also using a smaller backbone. \vspace{4.5mm}}
\label{table:fewshot_nobase}
\end{table}

\begin{table}[t]
\centering
\resizebox{1\linewidth}{!}{

\begin{tabular}{lccccc}
\toprule
\multirow{2}{*}{Method} & \multirow{2}{*}{Detector} & \multicolumn{4}{c}{COCO}\\
 & & 1\% & 2\% & 5\% & 10\% \\
\midrule\midrule
Supervised & \multirow{4}{*}{DDETR} &  11.31 $\pm$ 0.3 &  15.22 $\pm$ 0.32 &  21.33 $\pm$  0.2 &  26.34 $\pm$ 0.1 \\
SwAV & &  11.79 $\pm$ 0.3  &  16.02 $\pm$ 0.4 &  22.81 $\pm$ 0.3 &  27.79 $\pm$ 0.2 \\
ReSim & &  11.07 $\pm$ 0.4  & 15.26 $\pm$ 0.26 & 21.48 $\pm$ 0.1 &  26.56 $\pm$ 0.3
\\
\textbf{\model} &  & \textbf{ 14.58 $\pm$ 0.3} & \textbf{ 18.69 $\pm$ 0.2}    & \textbf{24.80 $\pm$ 0.2} & \textbf{ 29.12 $\pm$ 0.2} \\
\bottomrule                            
\end{tabular}
}
\caption{\textbf{Object detection using k\% of the labeled data on COCO.} The models are trained on \texttt{train2017} using k\% and then evaluated on \texttt{val2017}.}
\label{tab:semi}
\end{table}

\subsection{Semi-supervised Learning}
These experiments test how \model compares to semi-supervised methods, where small amounts of labeled data and large amounts of unlabeled data are used during training.

\noindent\textbf{Pretraining.} We pretrain \model (\defdetr) for $50$ epochs on MS~COCO \textit{train2017} without labels. 
\vspace{0.1cm}
\\
\noindent\textbf{Baselines.} We compare \model with a 
\defdetr model initialized with a supervised backbone from \imagenetthousand pretraining, which is the most direct baseline as all experiments are carried out on the same architecture and training data. We consider recent approaches for pretraining ResNet50 backbone for object detection like ReSim~\cite{xiao2021region} and SwAV~\cite{caron2020unsupervised}, for each we use the publicly released checkpoint.
\vspace{0.1cm}
\\
\noindent\textbf{Experiments.} We finetuned \model on random $k\%$ of \texttt{train2017} data for $k \in \{1, 2, 5, 10\}$, until convergence (validation performance stopped improving). In each setting, we train $5$ different models with different random seeds and report the mean and standard deviation.
\vspace{0.1cm}
\\
\noindent\textbf{Results.} \Cref{tab:semi} shows that \model outperforms existing pretraining methods, including a consistent improvement over the supervised pretraining baseline. We include a more broad comparison in the Supplementary~\Cref{tab:full_semi}, where we also compare to approaches that leverage both the labeled and unlabeled data via auxiliary losses~\cite{jeong2019consistency,sohn2020simple, liu2021unbiased, xu2021end}.

\begin{figure*}[t]
\centering
\begin{minipage}[t]{0.07\linewidth}
\vspace{-0.57in}
\centering
$\norm{\frac{\partial x}{\partial I}}$
\end{minipage}%
\includegraphics[width=0.22\textwidth, height=1in]{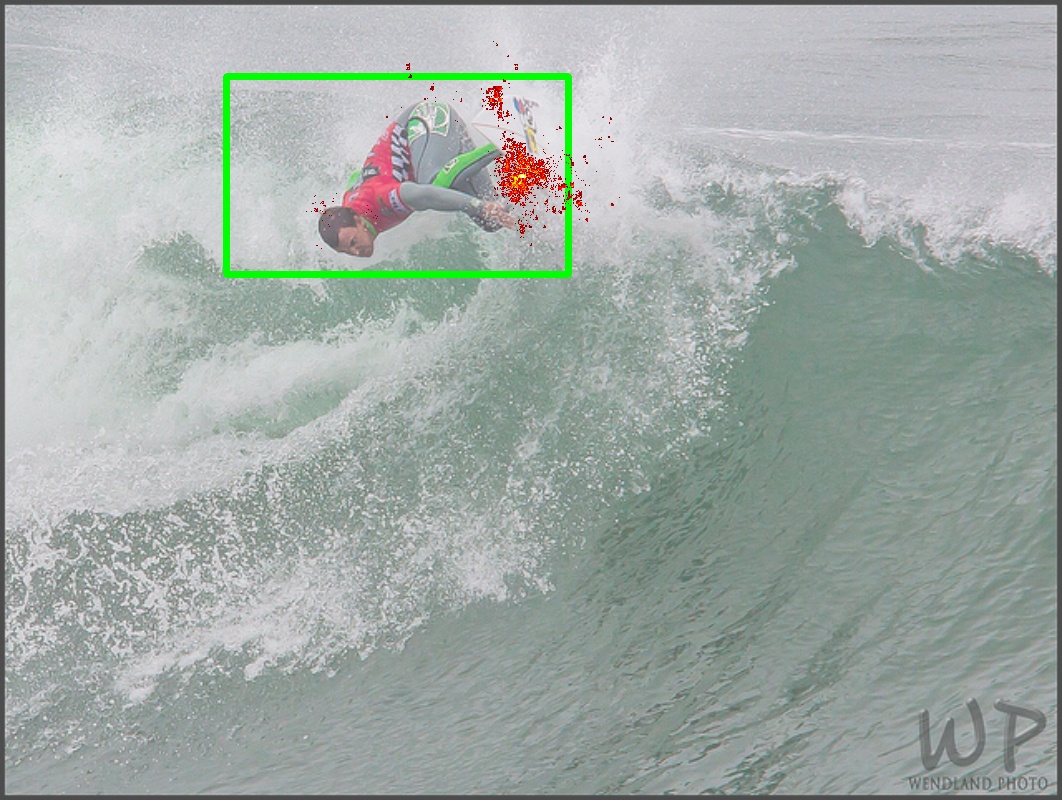}
\hfill
\includegraphics[width=0.22\textwidth, height=1in]{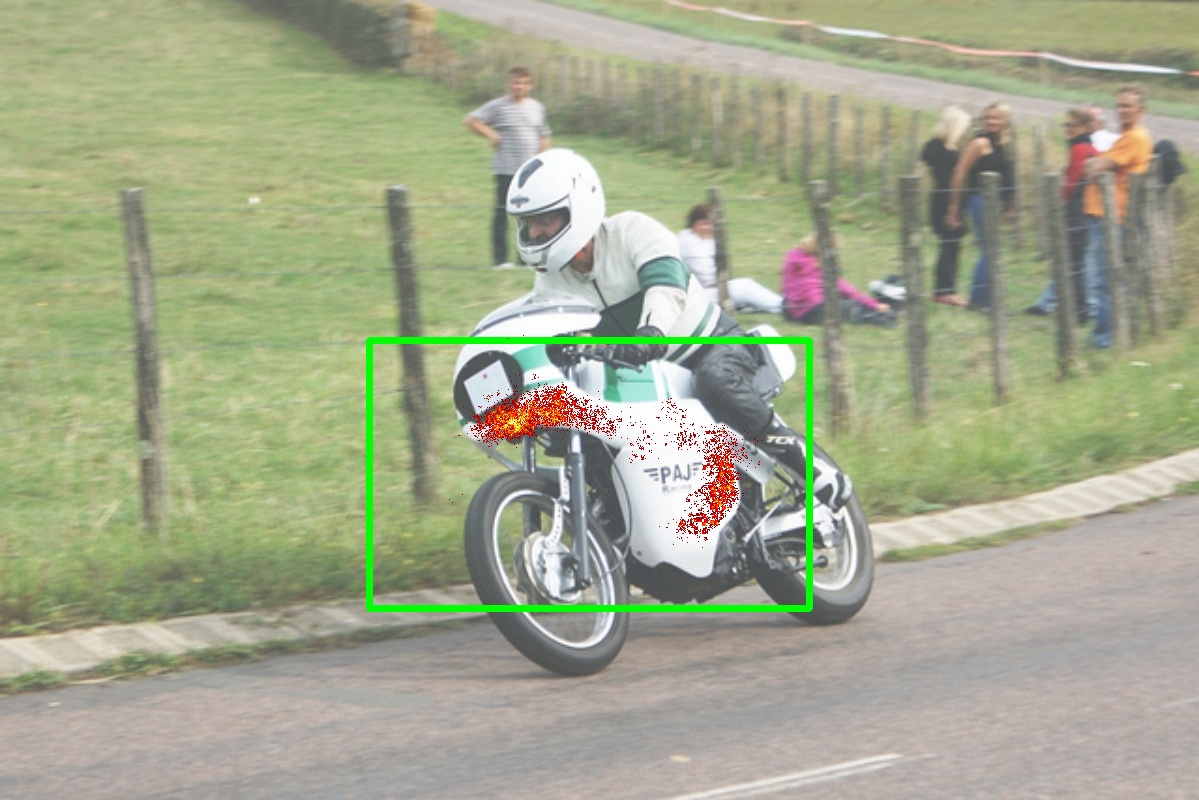}
\hfill
\includegraphics[width=0.22\textwidth, height=1in]{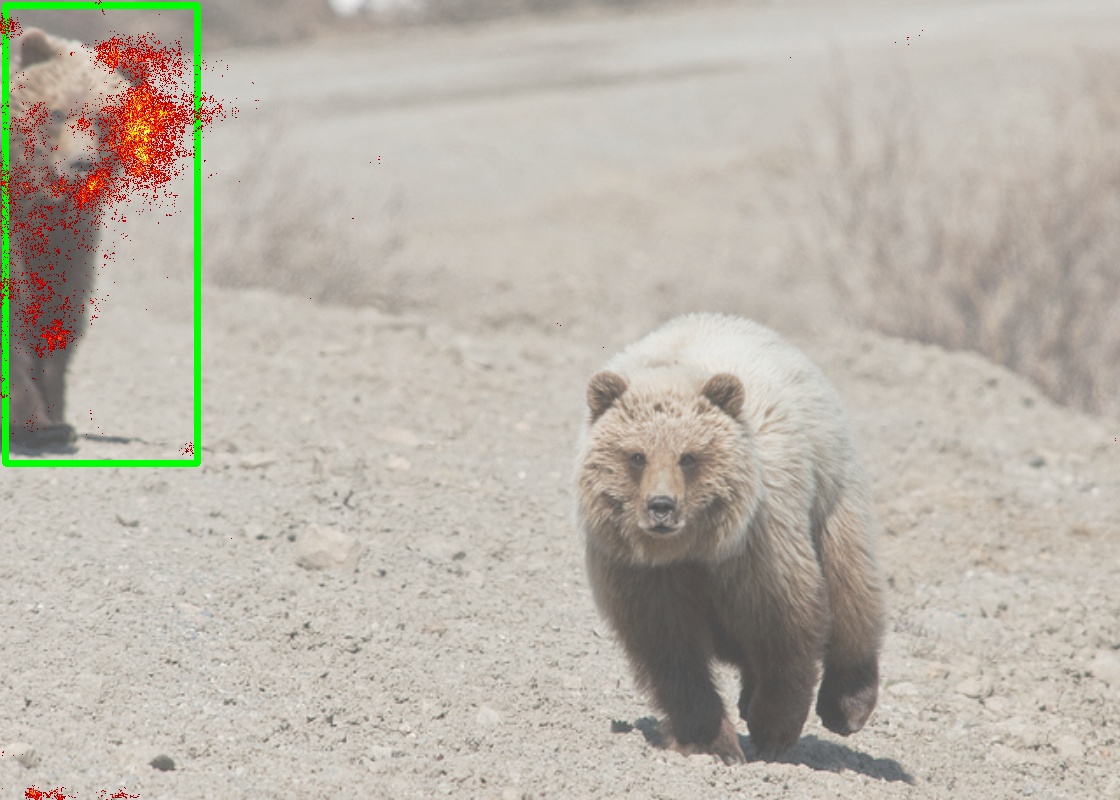}
\hfill
\includegraphics[width=0.22\textwidth, height=1in]{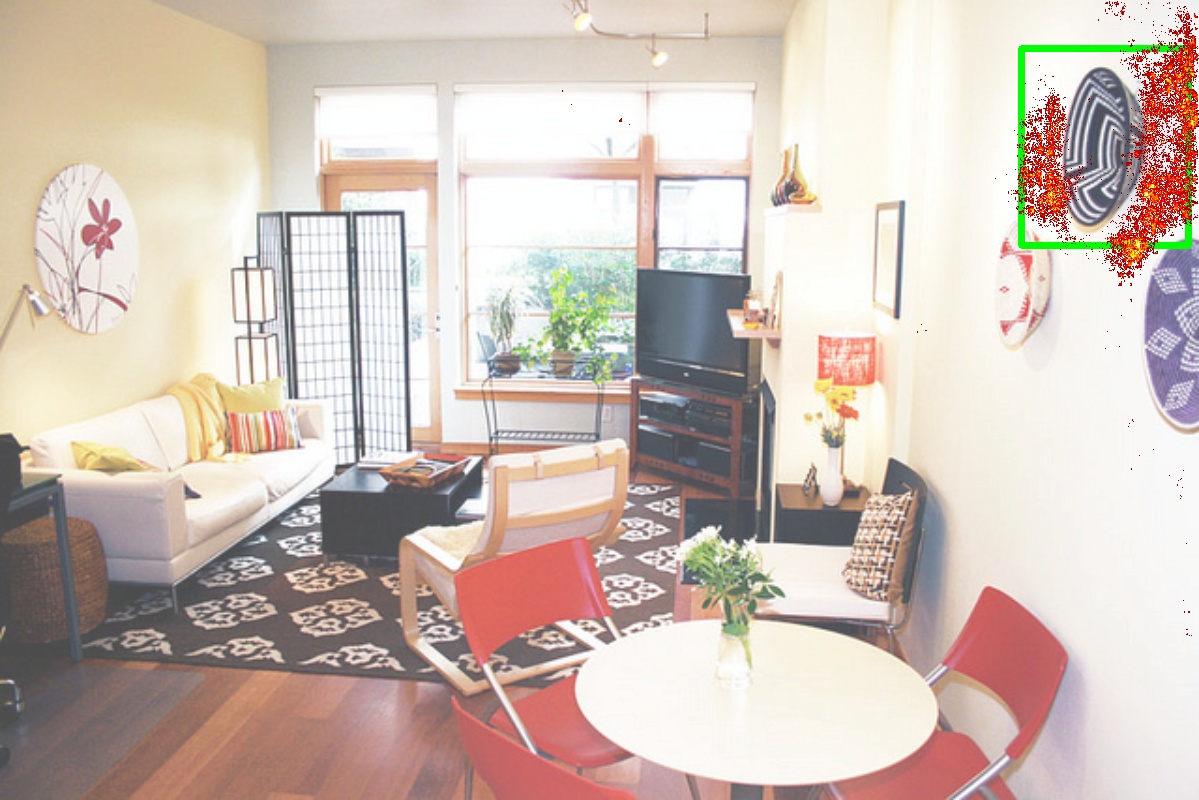}

\begin{minipage}[t]{0.07\linewidth}
\vspace{-0.57in}
\centering
$\norm{\frac{\partial y}{\partial I}}$
\end{minipage}%
\includegraphics[width=0.22\textwidth, height=1in]{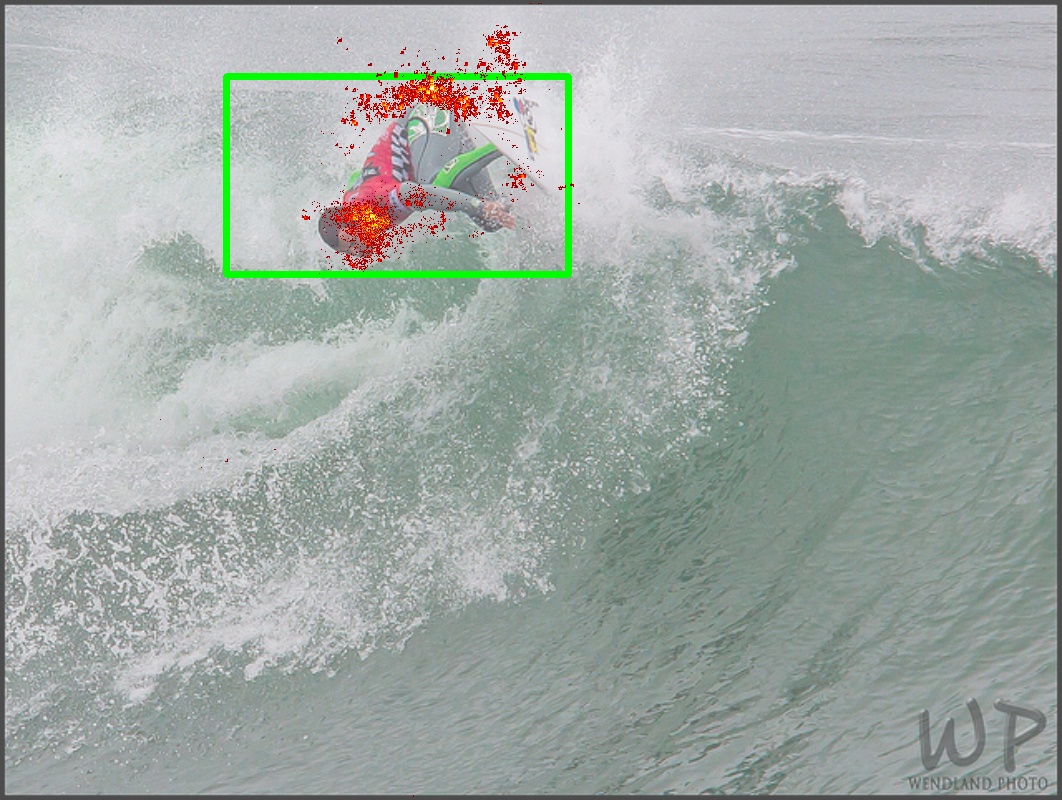}
\hfill
\includegraphics[width=0.22\textwidth, height=1in]{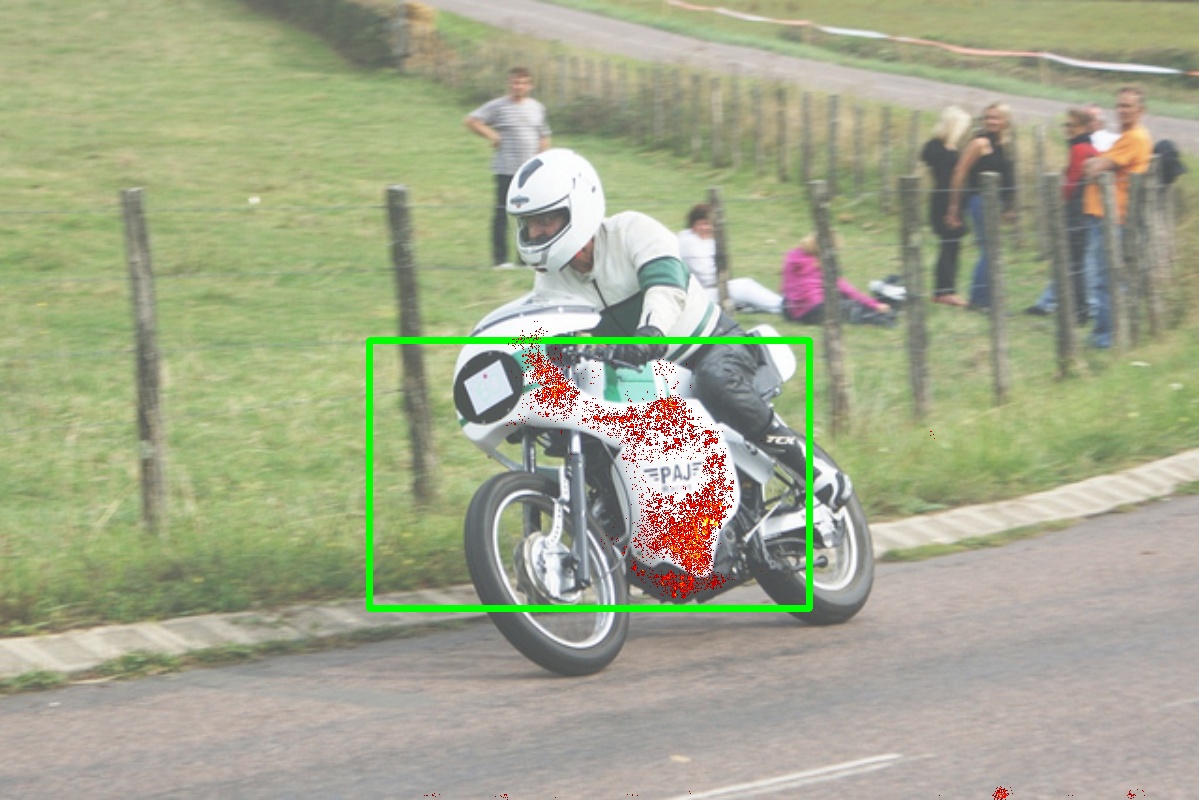}
\hfill
\includegraphics[width=0.22\textwidth, height=1in]{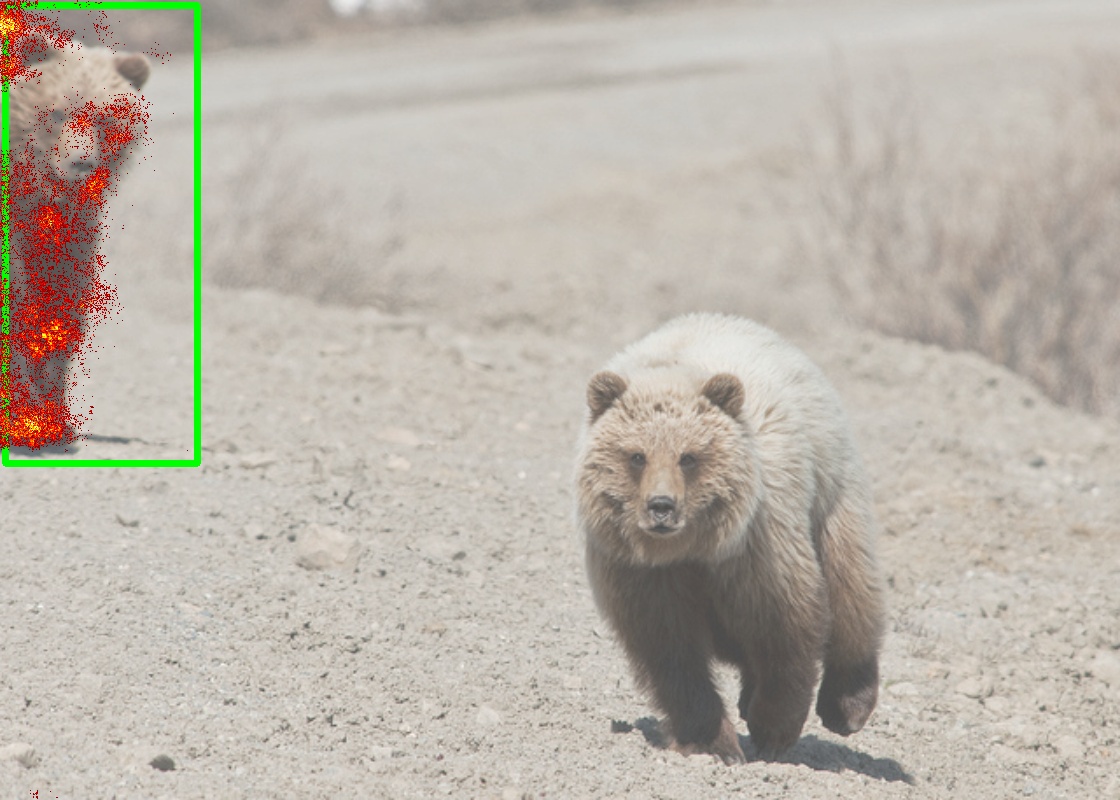}
\hfill
\includegraphics[width=0.22\textwidth, height=1in]{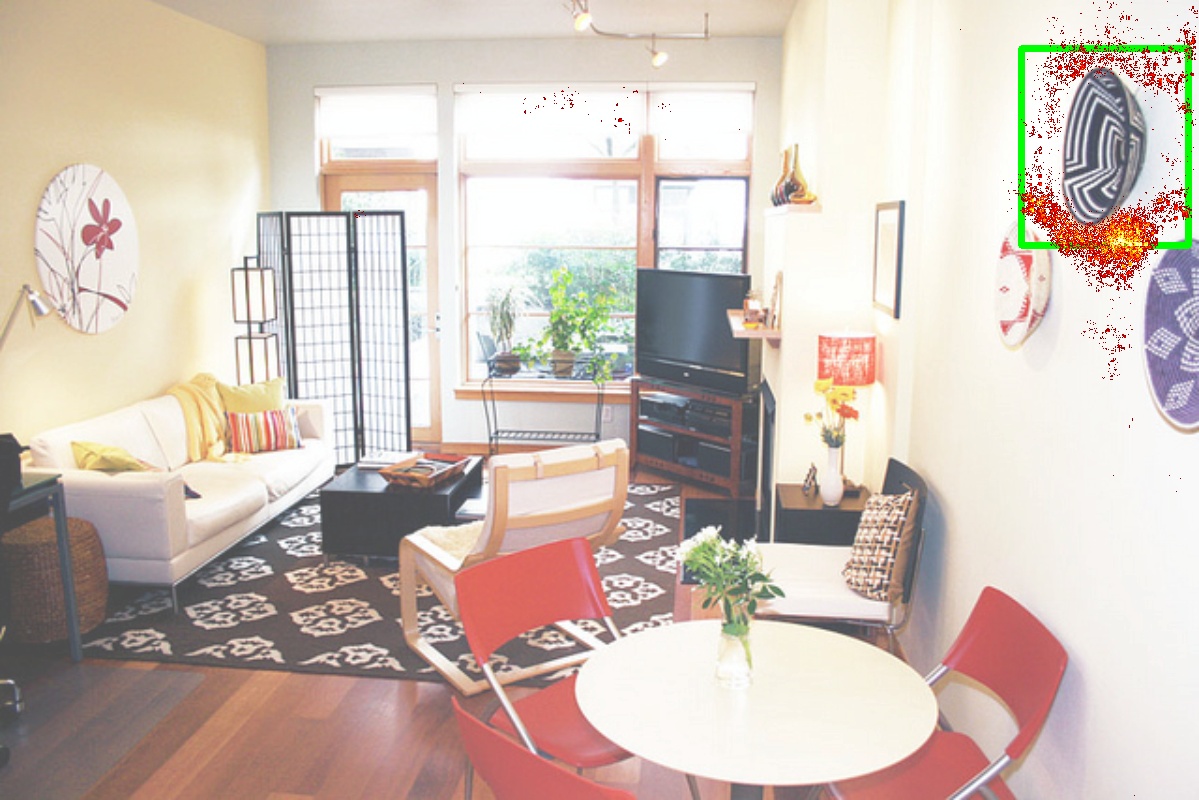}

\begin{minipage}[t]{0.07\linewidth}
\vspace{-0.57in}
\centering
$\norm{\frac{\partial z}{\partial I}}$
\end{minipage}%
\includegraphics[width=0.22\textwidth, height=1in]{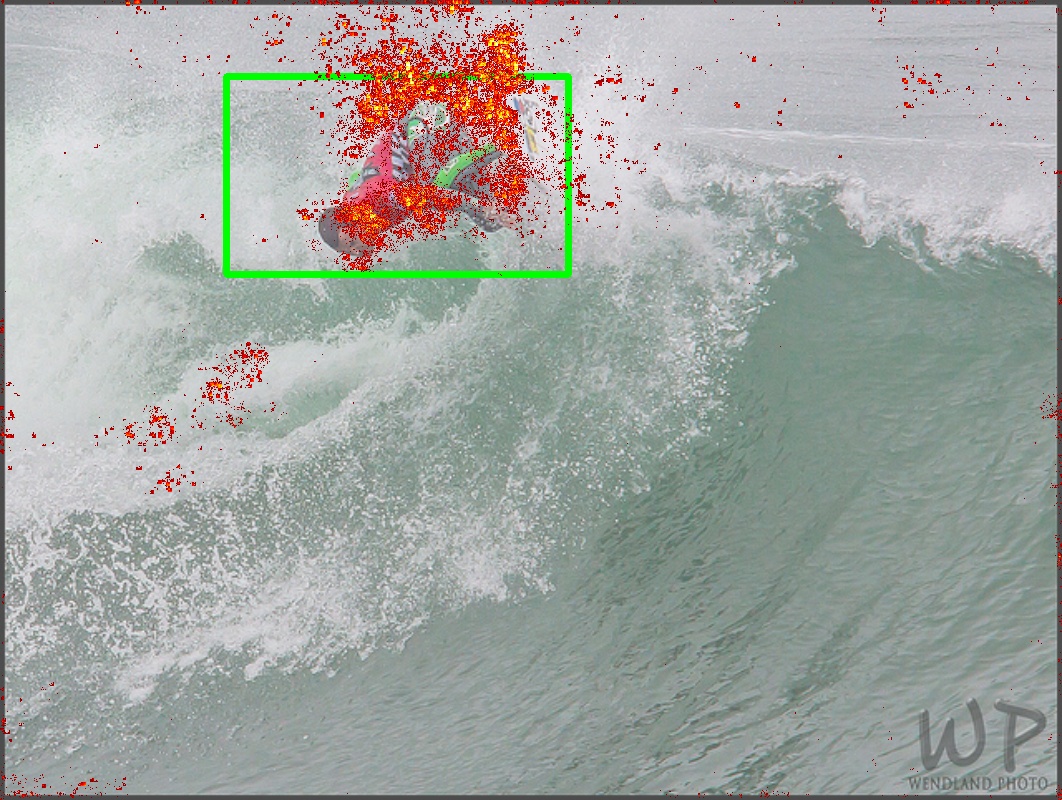}
\hfill
\includegraphics[width=0.22\textwidth, height=1in]{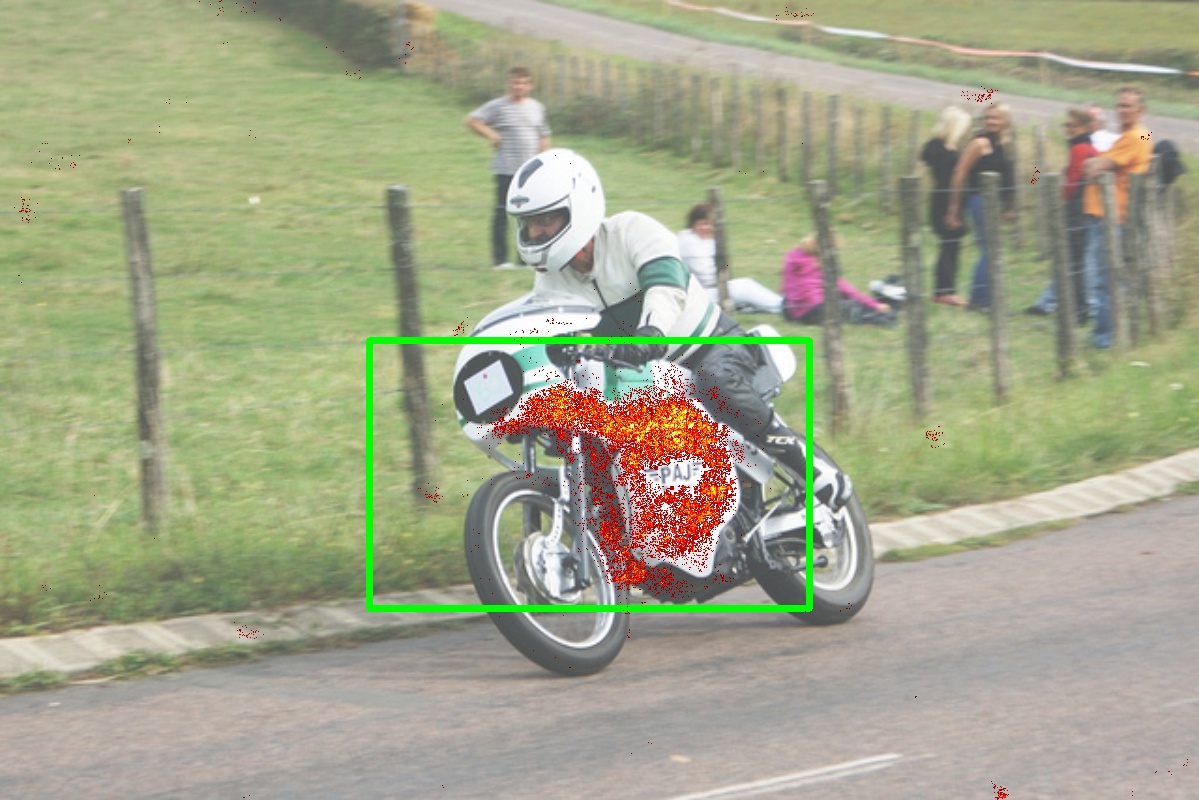}
\hfill
\includegraphics[width=0.22\textwidth, height=1in]{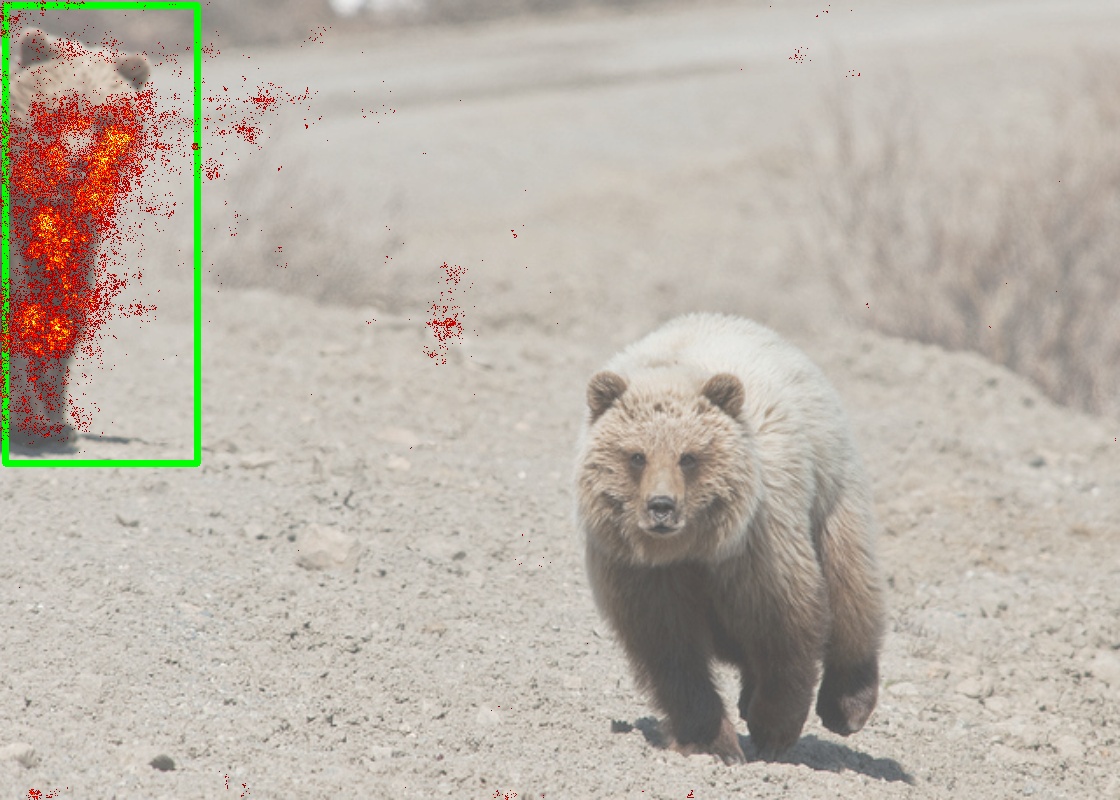}
\hfill
\includegraphics[width=0.22\textwidth, height=1in]{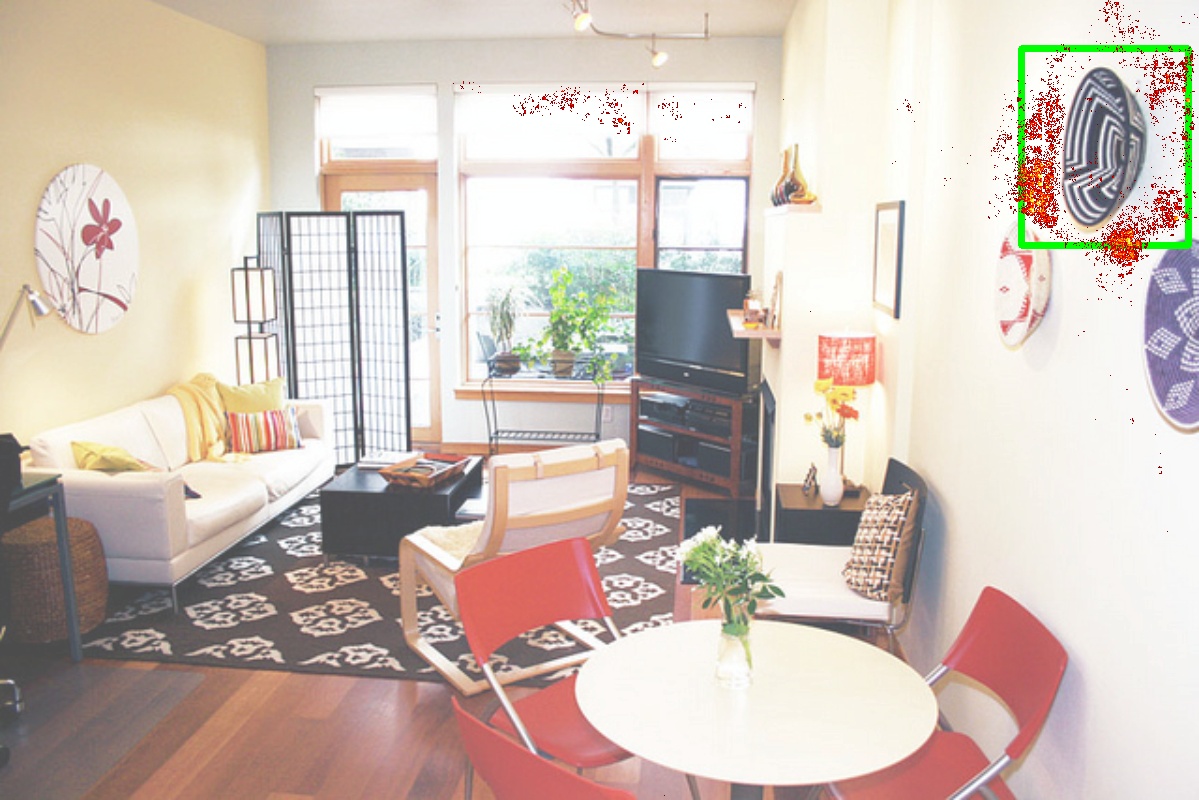}
\caption{\textbf{\model visualization.} We show the gradient norms from the unsupervised DETReg detection with respect to the input image $I$ for \textbf{(top)} the $x$ coordinate of the object center, \textbf{(middle)} the $y$ coordinate of the object center, \textbf{(bottom)} the feature-space embedding, $z$.}
\label{fig:qualitative}
\end{figure*}

\subsection{\model Analysis} \label{sec:visualize}
This section further explores and justifies the architectural and algorithmic choices used in the main experiments.
\label{ss:analysis}

\begin{table}[t]
\centering
\small

\resizebox{1\linewidth}{!}{
    \begin{tabular}{ccc|ccc}
    \toprule
    Proposals &  $L_{emb}$ & Frozen BB & $L_\text{class}$ $\downarrow$ & $L_\text{box}$ $\downarrow$ & AP 
    \\
    \midrule\midrule
    Shuffle & $\lambda_{e}=0$ & & 11.3 & .044 & 32.0 
    \\
    Top-K & $\lambda_{e}=0$ &  & 9.50 & .037 &  43.3 
    \\
    Top-K & $\lambda_{e}=1$ &  & 8.81 & .037 & 45.1 
    \\ 
    Top-K & $\lambda_{e}=2$ &  & 9.14 & .039 & 43.8 
    \\
    Top-K & $\lambda_{e}=1$ &  \checkmark & \textbf{8.61} & \textbf{.037} & \textbf{45.4} 
    \\
\bottomrule
\end{tabular}}
\caption{\textbf{Ablation studies.} This tables ablates region proposal sampling strategies, values of $\lambda_\text{emb}$, and whether to freeze backbones with \model trained on \textit{\imagenethundred} and finetuned on MS~COCO. 
Shuffling the region proposals across images led to a $11.3$ AP drop, $L_{emb}$ has a consistent performance, and freezing the backbone does not significantly change the performance.}
\label{table:ablations}
\end{table}

\begin{table}[t]
\centering
\small
\resizebox{1\linewidth}{!}{
    \begin{tabular}{lccccccc}
    \toprule
    Method & AP & AP$_\text{50}$ & AP$_\text{75}$ & $R@1$ & $R@10$ & $R@100$\\
    \midrule\midrule
    UP-DETR~\cite{dai2020up} & 0.0 & 0.0 & 0.0 & 0.0 & 0.0 & 0.4\\
    Rand. Prop. & 0.0 & 0.0 & 0.0 & 0.0 & 0.0 & 0.8\\
    Selective Search~\cite{uijlings2013selective} & 0.2 & 0.5 & 0.1 & 0.2& 1.5 & 10.9 \\
    \midrule
    {ImpSamp (ours)} & 0.7 & 2.0 & 0.1 & 0.3 & 1.8 & 9.0 \\
    {Random-K (ours)} & 0.7 & 2.4 & 0.2 & 0.5 & 2.9 & 11.7\\
    {Top-K (ours)} & \textbf{1.0} & \textbf{3.1} & \textbf{0.6} & \textbf{0.6} & \textbf{3.6} & \textbf{12.7} \\
    \bottomrule
    \end{tabular}
}
\caption[hypcap=false]{\textbf{Class agnostic object proposal evaluation on MS~COCO \texttt{val2017}}. The models are trained on IN100 and for each method, we consider the top 100 proposals. We show \model identifies objects more effectively than the previous methods.}
\label{table:unsupervised}
\end{table}

\noindent\textbf{Design Ablations.} \Cref{table:ablations} examines the contribution of the object localization and object embedding tasks in \model. 
To quantify the importance of using object-centric region proposals, we train \model while randomly shuffling the proposal box locations across images, as indicated by ``Shuffle'' in the ``Proposals'' column. Second, to assess the contribution of the embedding loss $L_{emb}$, we evaluate \model with different coefficients $\lambda_{e}\in\{0,1,2\}$. Finally, we validate that performance does not drop when freezing the backbone during training, i.e. that the performance benefits stem from the core \model contributions. All models are trained on \imagenethundred for $50$ epochs and finetuned on MS~COCO. 

\Cref{table:ablations} justifies our design choices: shuffling the region proposals across images led to a $11.3$ AP drop indicating that the object-centric proposal are important. We further see that the embedding loss $L_{emb}$ has a relatively consistent performance improvements with changes of $\leq 2$ AP for all setting, and we select $\lambda_e=1$ based on these results. Finally, the performance of \model with and without freezing the backbone encoder is relatively consistent with changes of $0.3$ AP points between the two settings.
\vspace{0.1cm}
\\
\noindent\textbf{Class Agnostic Object Detection.} We examine the class agnostic performance of \model variants discussed in~\Cref{sec:model}, as well as region proposal and pretraining approaches. The results reported in~\Cref{table:unsupervised} indicate that \model variants achieve improved performance over other pretraining approaches including solely using Selective Search. This indicates that coupling the object embedding and localization components in the \model model improves the localization ability. In addition, we observe that the Top-K region proposal selection strategy performs best in these ablations.
\vspace{0.1cm}
\\
\noindent\textbf{Robustness to different proposal methods.} We test how \model performs when pretrained with Selective Search proposals compared to Edge Box region proposals~\cite{zitnick2014edge}. Specifically, we pretrain \model on \imagenethundred and finetune on MS~COCO with 2\% and 10\% of random data. We find that both variants perform similarly well with AP of $21.8$ vs. $21.0$ for $2\%$ and a similar result of $36.2$ for $10\%$.
\vspace{0.1cm}
\\
\noindent\textbf{Visualizing \model.} \Cref{fig:qualitative} shows qualitative examples of \model unsupervised box predictions with \defdetr. Additionally, it shows the Saliency Map~\cite{simonyan2013deep} of the $x/y$ bounding box center and the object embedding with respect to the input image $I$. The first three columns show the attention focusing on the object edges for the $x/y$ predictions and $z$ for the predicted object embedding. The final column shows a case where the background plays a more important role than the object in the embedding. We believe this may be due to the CNN-based encoder focusing on the textures rather than the shapes in the region as discussed in~\cite{geirhos2018imagenet}, and we view further exploration of such characteristics as an intriguing direction for future work.

\section{Limitations}

\model's localization pretraining task uses simple region proposal methods for class-agnostic bounding-box supervision~\cite{uijlings2013selective, arbelaez2014multiscale, cheng2014bing, cheng2014global}. 
While \Cref{table:unsupervised} indicates that \model performance can improve beyond these methods, \model class-agnostic results remain far behind supervised counterparts. Furthermore, our experiments focused on DETR~\cite{carion2020end}-related architectures, but it may be possible that \model applies to more traditional detection architectures, which we leave for future work to explore. Finally, while \model improves training time, transformer-based object detectors still require significant computational resources to train.

\section{Conclusion}
We presented DETReg, an unsupervised pretraining approach for object detection with transformers using region priors. Through extensive empirical study, we showed \model learns representations in the unsupervised pretraining stage that lead to improvements in downstream performance for two different transformer models across three different datasets and many settings. We believe unsupervised pretraining holds the potential for positive social impact, mainly because it can utilize unlabeled data and reduce the need for massive amounts of labeled data which can be very expensive for fields like Medical Imaging. We do not anticipate a negative impact specific to our approach, but as with any model, we recommend careful validation before deployment.

\section*{Acknowledgements}
We would like to thank Sayna Ebrahimi and Jitendra Malik for their helpful feedback. This project has received funding from the European Research Council (ERC) under the European Unions Horizon 2020 research and innovation programme (grant ERC HOLI 819080). Prof. Darrell’s group was supported in part by DoD including DARPA's LwLL and/or SemaFor programs, as well as BAIR's industrial alliance programs. This work was completed in partial fulfillment for the Ph.D degree of the first author.

\clearpage

{\small
\bibliographystyle{ieee_fullname}
\bibliography{cvpr_2022}
}
\clearpage

\setcounter{section}{0}

\section*{Supplementary Material}
We start by providing the full implementation details of \model and include the complete PASCAL~VOC results. We then follow with additional analysis of \model pretraining as well as class agnostic performance and visualization.
\vspace{0.5mm}
\\
\noindent\textbf{Implementation Details.} 
Based on the ablations presented in \Cref{ss:analysis}, the default experiment settings are as follows. For region proposals, we compute Selective Search boxes online using the ``fast'' preset of the OpenCV implementation~\cite{opencv_library} and unless otherwise noted, we use the \model Top-K region selection variant (see \Cref{ss:object_localization_task}) and set $K=30$ proposals per-image.
We initialize the ResNet50 backbone of \model with SwAV~\cite{caron2020unsupervised}, which was pretrained with multi-crop views for 800 epochs on \imagenetthousand, and fix it throughout the pretraining stage. A similar SwAV encoder is used to encode region proposals, which are first cropped and resized to $128x128$.
In the object embedding branch, $f_{emb}$ and $f_{box}$ are MLPs with $2$ hidden layers of size $256$ followed by a ReLU~\cite{nair2010rectified} nonlinearity. The output sizes of $f_{emb}$ and $f_{box}$ are $512$ and $4$. $f_{cat}$ is implemented as a single fully-connected layer with $2$ outputs. 
We run the pretraining experiments using a batch size of $24$ per GPU on an NVIDIA DGX, V100 x8 GPUs machine, following the hyperparameter settings and image augmentations from existing works~\cite{zhu2020deformable, carion2020end}. Similarly, cropped regions are augmented before being fed to the encoder to obtain embeddings $\zz_i$. When finetuning, we drop the $f_{emb}$ branch, and set the size of the last fully-connected layer of $f_{cat}$ to be the number of classes in the target dataset plus a background class.

\begin{table}
\makeatletter\def\@captype{table}
\centering
\resizebox{0.8\linewidth}{!}{
\begin{tabular}{l|c|ccc}
\toprule
Method & Detector & AP & AP$_{50}$ & AP$_{75}$ \\
\midrule
\midrule
Supervised & \multirow{14}{*}{FRCN} &  56.1 & 82.6 & 62.7 \\ 
InsDis~\cite{wu2018unsupervised} & & 55.2 &80.9 &61.2 \\ 
Jigsaw~\cite{goyal2019scaling} & & {48.9} & {75.1} & {52.9}\\
NPID++~\cite{misra2020self} & & {52.3} & {79.1} & {56.9}  \\
SimCLR~\cite{chen2020simple} & &{51.5} & {79.4} & {55.6}  \\
PIRL~\cite{misra2020self} & &{54.0} & {80.7} & {59.7}\\
BoWNet~\cite{gidaris2020learning} & &{55.8} & {81.3} & {61.1}\\
MoCo~\cite{he2019momentum} & &{55.9} & {81.5} & {62.6}\\
MoCo-v2~\cite{chen2020improved} & &{57.0} & {82.4} & {63.6} \\
SwAV~\cite{caron2020unsupervised} & &{56.1} & {82.6} & {62.7} \\
DenseCL~\cite{wang2020dense} & &{58.7} & {82.8} & {65.2} \\
DetCo~\cite{xie2021detco} & &58.2 & 82.7 & 65.0 \\ 
ReSim~\cite{xiao2021region} & &{59.2} & {82.9} & 65.9 \\
\midrule
Supervised & \multirow{2}{*}{DETR} &  54.1  &  78.0 &  58.3 \\ 
UP-DETR~\cite{dai2020up} & &57.2 & 80.1 & 62.0 \\
\midrule
Supervised & \multirow{3}{*}{DDETR} & 59.5  &  82.6 &  65.6 \\
SwAV~\cite{caron2020unsupervised} &  & 61.0 & 83.0 & 68.1 \\
\model & & \textbf{63.5} & \textbf{83.3} & \textbf{70.3} \\
\bottomrule                        
\end{tabular}}
\caption{\textbf{Object detection finetuned on PASCAL~VOC.} The model is finetuned on PASCAL~VOC \texttt{trainval07+2012} and evaluated on \texttt{test07}. Models are based on Faster-RCNN~\cite{ren2015faster} (FRCN), DETR~\cite{carion2020end}, and~\defdetr~\cite{zhu2020deformable} (DDETR). Bold values indicate an improvement $\geq 0.3$ AP.}
\vspace{-5mm}
\label{tab:voc}
\end{table}

\subsection*{Object Detection in Full Data Regimes} We reported \model results on the PASCAL~VOC benchmark in~\Cref{subsec:fulldata}. Here we include the full table, containing more past pretraining approaches using three different object detectors (see~\Cref{tab:voc}). We observe that using the Deformable-DETR detector, the supervised pretraining baseline is superior to past pretraining approaches and that \model pretraining improves over it by 4 points (AP).

\begin{figure}
 \centering
     \centering
     \includegraphics[width=0.47\textwidth]{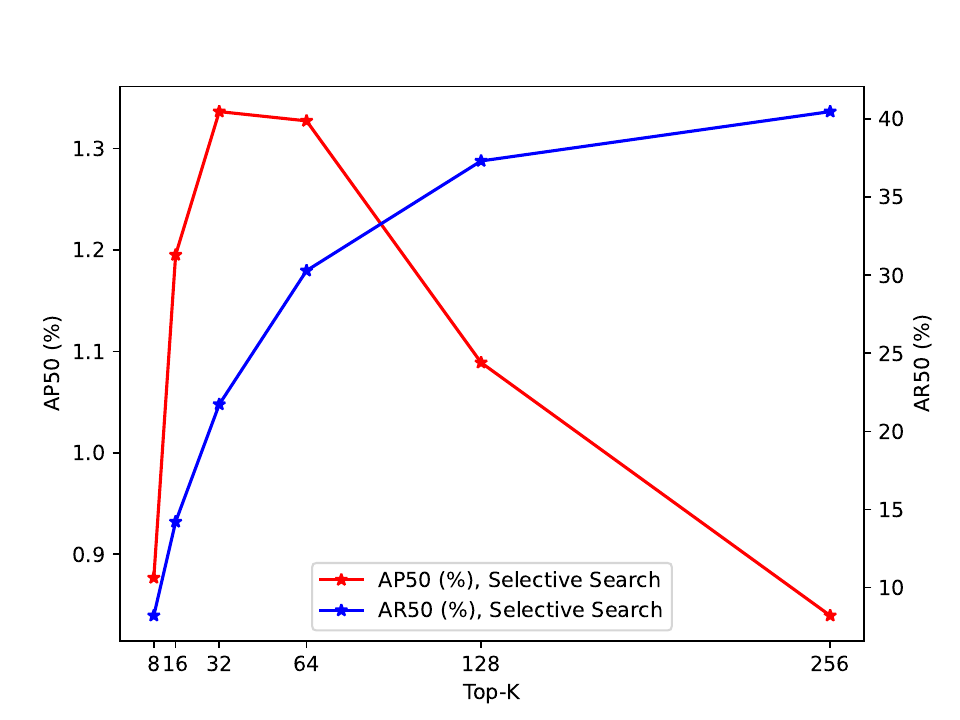}
     \caption{\textbf{Top-K proposals performance of Selective Search.} Using different values of $K$, w evaluate the class agnostic performance of Selective Search on MS~COCO 2017 validation split.}
    \label{fig:ar_ap}
    \vspace{-4mm}
\end{figure}

\subsection*{Semi-supervised Learning} We reported \model results and comparisons to other pretraining approaches like~\cite{caron2020unsupervised, xiao2021region} when using limited amounts of data. In~\Cref{tab:full_semi}, we include comparisons to semi-supervised works~\cite{jeong2019consistency,sohn2020simple, liu2021unbiased, xu2021end} that leverage both the labeled and unlabeled data in training via auxiliary losses.

\begin{table*}[t]
\centering
\resizebox{0.82\linewidth}{!}{

\begin{tabular}{lcccccccc}
\toprule
\multirow{2}{*}{Method} & \multirow{2}{*}{Approach} & \multirow{2}{*}{Detector}  & \multicolumn{4}{c}{COCO}\\
 & & & 1\% & 2\% & 5\% & 10\% \\
\midrule\midrule
CSD~\cite{jeong2019consistency} & \multirow{4}{*}{Auxiliary} & \multirow{4}{*}{FRCN} & 10.5 $\pm$ 0.1  & 13.9 $\pm$ 0.1 & 18.6 $\pm$ 0.1 & 22.5 $\pm$ 0.1  \\ 
STAC~\cite{sohn2020simple} & & & 14.0 $\pm$ 0.6 & 18.3 $\pm$ 0.3 & 24.4 $\pm$ 0.1 & 28.6 $\pm$ 0.2 \\ 
U-T~\cite{liu2021unbiased} & & & 20.8 $\pm$ 0.1  & 24.3 $\pm$ 0.1 & 28.3 $\pm$ 0.1 & {31.5 $\pm$ 0.1} \\
S-T~\cite{xu2021end} & & & \textbf{20.5 $\pm$ 0.4}   &  $-$ & \textbf{30.7 $\pm$ 0.1} & {\textbf{34.0 $\pm$ 0.1}} \\
\midrule
Supervised & \multirow{4}{*}{Pretraining} & \multirow{4}{*}{DDETR} &  11.31 $\pm$ 0.3 &  15.22 $\pm$ 0.32 &  21.33 $\pm$  0.2 &  26.34 $\pm$ 0.1 \\
SwAV & & &  11.79 $\pm$ 0.3  &  16.02 $\pm$ 0.4 &  22.81 $\pm$ 0.3 &  27.79 $\pm$ 0.2 \\
ReSim & & &  11.07 $\pm$ 0.4  & 15.26 $\pm$ 0.26 & 21.48 $\pm$ 0.1 &  26.56 $\pm$ 0.3
\\
\textbf{\model} & &  & \textbf{ 14.58 $\pm$ 0.3} & \textbf{ 18.69 $\pm$ 0.2}    & \textbf{24.80 $\pm$ 0.2} & \textbf{ 29.12 $\pm$ 0.2} \\
\bottomrule                            
\end{tabular}
}
\caption{\textbf{Object detection using k\% of the labeled data on COCO.} The models are trained on \texttt{train2017} using k\% and then evaluated on \texttt{val2017}. Methods like~\cite{liu2021unbiased} utilize auxiliary losses during the training stage using unlabeled data, whereas \model utilizes unlabeled data during the pretraining stage only.}
\label{tab:full_semi}
\end{table*}

\begin{figure*}
 \centering
     \centering
     \includegraphics[width=1\textwidth]{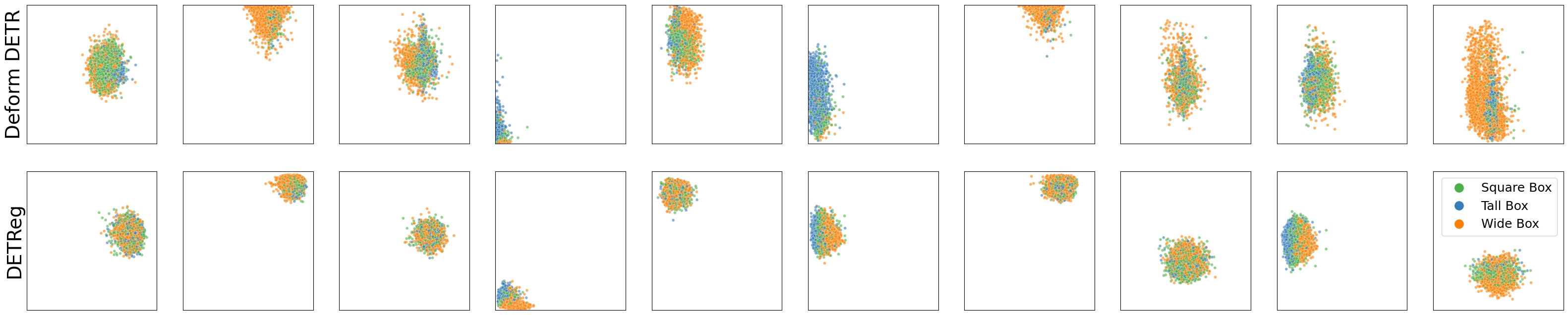}
     \caption{\textbf{\model slots specialize in specific areas in the image and uses a variety of box sizes much like Deformable DETR}. Each square corresponds to a DETR slot, and shows the location of its bounding box predictions. We compare 10 random slots of the supervised Deformable DETR \textbf{(top)} and unsupervised DETReg \textbf{(bottom)} decoder for the MS~COCO 2017 val dataset. Each point shows the center coordinate of the predicted bounding box, where following a similar plot in \cite{carion2020end}, a {\color{green}green point} represents a square bounding box, a {\color{orange}orange point} is a large horizontal bounding box, and a {\color{blue}blue point} is a large vertical bounding box. Deformable DETR has been trained on MS~COCO 2017 data, while DETReg has only been trained on unlabeled ImageNet data. Similar DETReg and Deformable DETR slots were manually chosen for illustration.}
    \label{fig:object_queries}
\end{figure*}

\begin{figure*}
 \centering
     \centering
     \includegraphics[width=1\textwidth]{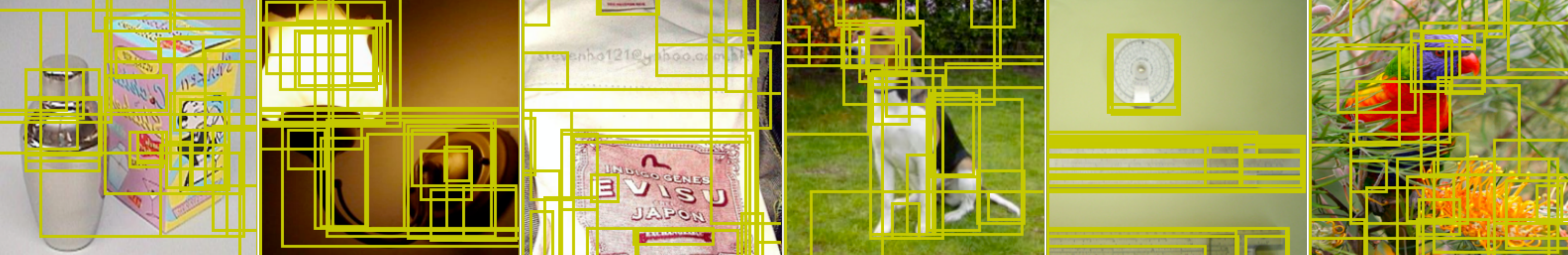}
     \caption{\textbf{TopK Selective Search proposals on ImageNet.} Using K=30, the proposals typically cover objects and parts-of-objects in the image.}
    \label{fig:topk_in}
\end{figure*}

\subsection*{DETReg Analysis} 
In~\Cref{sec:visualize} we analyzed \model, including the model ablations, class agnostic results, visualization and robustness. Here we further examine the pretrained \model model including the class agnostic results, and TopK selection policy.

\paragraph{Improved Encoder, improved \model.} We test how \model performs when object embeddings are obtained with different image encoders. Specifically, we pretrain \model on \imagenethundred using SwAV trained for 400 epochs compared to a superior variant trained for 800 epochs with multi-crops. We finetune on MS~COCO with 1\% data and observe the improved encoder achieves $1$ AP improvement (27.7 vs 26.7).
\paragraph{\model TopK selection policy.} Using Selective Search, we examine the class agnostic performance when using TopK policy. We report the precision and recall in~\Cref{fig:ar_ap}. In this paper, we have used $K=30$ (see~\Cref{fig:topk_in}), which emphasizes precision over recall. This might imply that \model performs well given high precision proposals.

\begin{figure*}
 \centering
     \centering
     \includegraphics[width=1\textwidth]{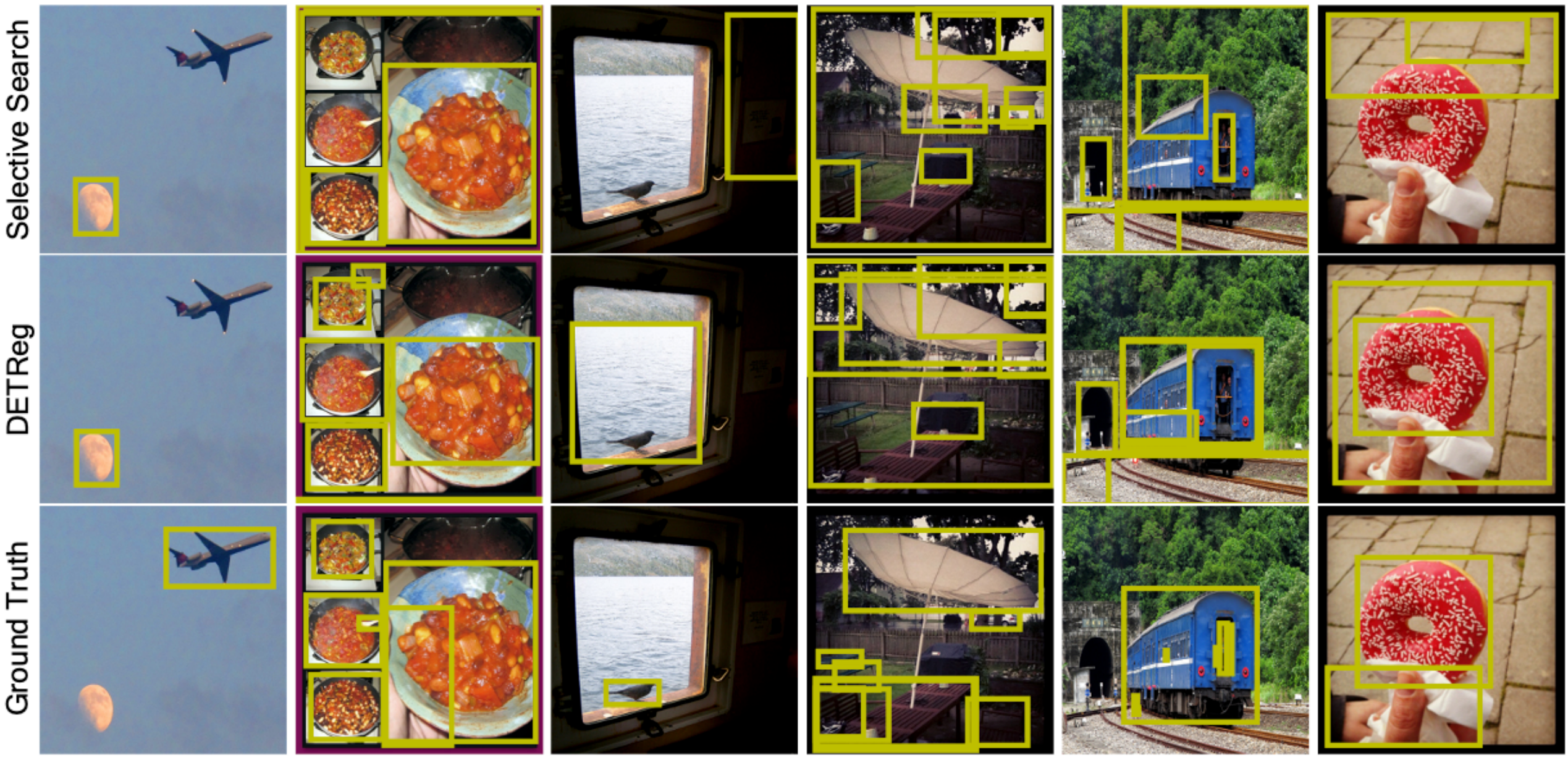}
     \caption{\textbf{Class Agnostic object detection visualization.} Examples predictions using Selective Search and \model on random MS~COCO images. For every image annotated with $M$ boxes, only the top $M$ predictions are shown.}
    \label{fig:viz_models}
\end{figure*}

\paragraph{\model Slots Visualization.} We examine the learned object queries slots (see~\Cref{fig:object_queries}) and observe they are similar to those in \defdetr, despite not using any human annotated data. Nevertheless, the \defdetr slots have greater variance with respect to locations and they tend to specialize more in particular boxes shapes.
\paragraph{Class Agnostic Object Detection.} The quantitative results in~\Cref{sec:visualize} indicate that \model improves over Selective Search. The included qualitative examples of \model on MS~COCO (see~\Cref{fig:viz_models}) supports a similar conclusion, indicating that \model outperforms Selective Search but still much behind the ground truth labeled data.

\end{document}